\RequirePackage{fix-cm}
\documentclass[italian,english,british]{paper}
\usepackage{utopia}
\usepackage[T1]{fontenc}
\usepackage[latin9]{inputenc}
\usepackage{listings}
\usepackage[a4paper]{geometry}
\usepackage{color}
\usepackage{babel}
\usepackage{array}
\usepackage{longtable}
\usepackage{prettyref}
\usepackage{textcomp}
\usepackage{ifsym}
\usepackage{url}
\usepackage{multirow}
\usepackage{amsthm}
\usepackage{amsmath}
\usepackage{fixltx2e}
\usepackage[numbers]{natbib}
\inputencoding{utf8}
\usepackage[unicode=true,
 bookmarks=true,bookmarksnumbered=true,bookmarksopen=false,
 breaklinks=false,pdfborder={0 0 1},backref=false,colorlinks=false]
 {hyperref}
\hypersetup{pdftitle={An Experiment on the Conncetion between the Description Logicsâ Family DL<âÏ0> and the Real World},
 pdfauthor={Antonio Pisasale},
 pdfsubject={SPARQL queries},
 pdfkeywords={Semantic Web, RDF, DL, Description Logics, SPARQL, Database}}
\inputencoding{latin9}

\makeatletter

\newcommand{\noun}[1]{\textsc{#1}}
\providecommand{\tabularnewline}{\\}

\usepackage{enumitem}		
\newlength{\lyxlabelwidth}      
\newenvironment{elabeling}[2][]%
{\settowidth{\lyxlabelwidth}{#2}
\begin{description}[font=\normalfont,style=sameline,
leftmargin=\lyxlabelwidth,#1]}
{\end{description}}
\newcommand{\code}[1]{\texttt{#1}}
\newcommand{\strong}[1]{\textbf{#1}}

\renewcommand{\citet}{\citep}

\usepackage{xspace}
\xspaceaddexceptions{]\}-<>()}

\newcommand{\allpiz}{\ensuremath{\mathcal{\forall}_0^\pi}}
\newcommand{\mlsscart}{\ensuremath{\mathsf{MLSS}_{2,m}^{\times}}\space}
\newcommand{\dl}{\ensuremath{\mathcal{DL}}\xspace}
\newcommand{\sroiqd}{\ensuremath {s\mathcal{ROIQ}_{(D)}}\space}
\newcommand{\SxroiqD}{\ensuremath{\mathcal{SROIQ}(D)}\space}
\newcommand{\shifd}{\ensuremath{\mathcal{SHIF}_{(D)}}\xspace}
\newcommand{\shoind}{\ensuremath{\mathcal{SHOIN}_{(D)}}}
\newcommand{\I}{\space\ensuremath{\mathcal{I}}\xspace}
\newcommand{\deltaI}{\ensuremath{\Delta^\I}\xspace}
\newcommand{\deltaII}{\ensuremath{\deltaI\times\deltaI}\xspace}
\newcommand{\ifunc}{\ensuremath{^\I}\xspace}
\newcommand{\aI}{\ensuremath{a^\I}\xspace}
\newcommand{\abI}{\ensuremath{\left(a,b\right)^\I}\xspace}
\newcommand{\AI}{\ensuremath{A^\I}\xspace}
\newcommand{\CI}{\ensuremath{C^\I}\xspace}
\newcommand{\DI}{\ensuremath{D^\I}\xspace}
\newcommand{\PI}{\ensuremath{P^\I}\xspace}
\newcommand{\RI}{\ensuremath{R^\I}\xspace}
\newcommand{\SI}{\ensuremath{S^\I}\xspace}
\newcommand{\K}{\space\ensuremath{\mathcal{K}}\space}
\newcommand{\Nu}{\space\ensuremath{\mathcal{N}}\space}
\newcommand{\Qu}{\space\ensuremath{\mathcal{Q}}\space}
\newcommand{\Oo}{\space\ensuremath{\mathcal{O}}\space}
\newcommand{\Fu}{\space\ensuremath{\mathcal{F}}\space}

\newcommand{\Hi}{\space\ensuremath{\mathcal{H}}\space}
\newcommand{\Sx}{\space\ensuremath{\mathcal{S}}\space}
\newcommand{\Ro}{\space\ensuremath{\mathcal{R}}\space}
\newcommand{\alc}{\ensuremath{\mathcal{ALC}}\space}
\newcommand{\supm}{\ensuremath{^-}\space}
\newcommand{\supp}{\ensuremath{^+}\space}
\newcommand{\sups}{\ensuremath{^*}\space}
\newcommand{\ABox}{\textsf{ABox}\space}
\newcommand{\TBox}{\textsf{TBox}\space}
\newcommand{\OWLLite}{\textsf{OWL}\space\textsf{Lite}\xspace}
\newcommand{\OWLDL}{\textsf{OWL}\space\textsf{DL}\xspace}
\newcommand{\OWLFull}{\textsf{OWL}\space\textsf{Full}\xspace}
\newcommand{\OWLoneDL}{\textsf{OWL}\space\textsf{1}\space\textsf{DL}\xspace}
\newcommand{\OWLoneLite}{\textsf{OWL}\space\textsf{1}\space\textsf{Lite}\xspace}
\newcommand{\OWLoneFull}{\textsf{OWL}\space\textsf{1}\space\textsf{Full}\xspace}
\newcommand{\OWLtwoDL}{\textsf{OWL}\space\textsf{2}\space\textsf{DL}\xspace}
\newcommand{\OWLtwoFull}{\textsf{OWL}\space\textsf{2}\space\textsf{Full}\xspace}
\newcommand{\OWLtwoEL}{\textsf{OWL}\space\textsf{2}\space\textsf{EL}\xspace}
\newcommand{\OWLtwoQL}{\textsf{OWL}\space\textsf{2}\space\textsf{QL}\xspace}
\newcommand{\OWLtwoRL}{\textsf{OWL}\space\textsf{2}\space\textsf{RL}\xspace}

\usepackage[bottom,multiple]{footmisc}

\AtBeginDocument{

}

\makeatother

\begin{document}

\author{\noindent \textrm{\large Domenico Cantone }\textrm{\normalsize (\code{\noindent cantone@dmi.unict.it})}\textrm{\large }\\
\textrm{\large Antonio Pisasale }\textrm{\normalsize (\code{\noindent fluoro@email.it})}}

\title{\noindent \textrm{\noun{\Huge An Experiment on the Connection between
the Description Logics' Family}}\linebreak{}
\textrm{\Huge \dl<\allpiz> }\textrm{\noun{\Huge and the Real World}}}

\maketitle
~\thispagestyle{empty}
\begin{abstract}
This paper describes the analysis of a selected testbed of Semantic
Web ontologies, by a SPARQL query, which determines those ontologies
that can be related to the description logic \dl<\allpiz>, introduced
in \citet{C+11} and studied in \citet{L12}. We will see that a reasonable
number of them is expressible within such computationally efficient
language. We expect that, in a long-term view, a temporalization of
description logics, and consequently, of OWL(2), can open new perspectives
for the inclusion in this language of a greater number of ontologies
of the testbed and, hopefully, of the ``real world''.
\end{abstract}

\section*{\noun{INTRODUCTION}}

In the last years, Semantic Web has increasingly expanded its area
of influence. Being an innovative instrument for the retrieval of
not expressly stored information and a way of organizing concepts
and relations by their meaning, it broadens up plenty of horizons
for knowledge representation. Though, only a bunch of experts and
researchers know that under a suprisingly vast dimension of new features
there lies a mathematical and logical structure inside which they
fight day by day for the balancing of expressiveness and efficiency.
The \emph{Description Logics} formalisms, which we will see in Section
\ref{sec:THEORETICAL-FUNDAMENTALS}, are the formal bases for the
so-called Web 3.0, that should allow one to automatically infer (and
retrieve) new information from \emph{reasoning} on ``sematicized''
knowledge repositories. Some examples of families of logics are described
in the following, together with a short coverage of arguments such
as RDF graphs and OWL.

Section \ref{sec:ANALYSIS-AND-RESULTS} introduces the description
logic \dl<\allpiz>, showing notable characteristics of expressive
power and polynomial complexity. In the same section, a more detailed
description of the analysis, which may be seen as a conceptual experiment,
follows. Our purpose is to show that a good number of real-world ontologies
may be related to this family of logics. Results are promising enough
to spur us to stay on this path and complement it with studies on
temporalization of Description Logics and, consequently, of Semantic
Web (briefly touched in this report), which we will approach in the
near future.

The SPARQL query which was specifically created to assess the membership
of some real-world ontologies to the aforementioned logic is reported
in Appendix \ref{sec:app-a}.

\section{\noindent \textrm{\noun{\label{sec:THEORETICAL-FUNDAMENTALS}THEORETICAL
FUNDAMENTALS}}}

\subsection{Description Logics}

\selectlanguage{italian}%
\thispagestyle{empty}\foreignlanguage{british}{\hspace*{0.7cm}\emph{Description
Logics} (\emph{DL}s) are a family of formalisms for knowledge representation,
built on logic-based semantics. They are founded on some fundamental
ideas:}
\selectlanguage{british}%
\begin{itemize}
\item basic syntactic ``blocks'' are atomic \strong{concepts} (1-ary
predicates), atomic \strong{roles} (2-ary predicates) and \strong{individuals}
(costants);
\item the expressiveness of a certain language depends on the use of a set
of chosen constructors, that give birth to complex concepts and roles
starting from existing ones;
\item by means of classification of concepts, a \emph{subsumption }hierarchy
is established, which specifies what concept includes or is included
by another;
\item implied knowledge is automatically obtained through a logic procedure,
called \emph{reasoning}, essentially based on the application of inference
rules to subsumption between concepts, and an instance definitions
between individuals and the latter.
\end{itemize}
On the assumption that a Knowledge Representation System (\emph{KR-System})
is to give an answer to a user query, \emph{reasoning} algorithms
for DLs should be regarded as decisional procedures which return a
positive or negative verdict. This raises the decision problem for
such languages. Furthermore, having an answer does not always mean
getting it in a reasonable lapse of time, and that compels us to consider
also the complexity of algorithms at stake. Decidability and complexity
directly depend on the expressive power of the description logic we
use: whereas very expressive DLs tend to have inference problems and
be computationally \emph{hard} (or even undecidable), the DLs that
are more efficient in \emph{reasoning} turn out to be not espressive
enough to represent all concepts and relations in the domain of interest.
Research still going on in the field of DLs just aims at the ability
of balancing expressiveness against efficiency, while not dropping
semantic precision that could make it applicable to real world situations.

A \emph{Description Logics Knowledge Base} (\emph{DLKB}) is made of
two components: \TBox (\emph{Terminological Box}) and \ABox (\emph{Assertional
Box}). The former contains the vocabulary, i.e., the definitions of
atomic and non-atomic concepts and roles, called \emph{axioms}, whereas
the latter describes individuals in terms of this vocabulary, i.e.
it includes the declarations, called \emph{assertions}, of their instances.
By means of the TBox we can name complex descriptions of concepts
and roles. The language for such a naming is what distinguishes one
DL from another and is based on a \emph{model-theoretic} semantics.
Thus statements in TBox and ABox can be regarded as first-order logic
formulas. \emph{Reasoning} procedures for the terminological part
are used to verify the \emph{satisfiability} of a description (i.e.,
its non-contradictoriness), or whether it may be \emph{subsumed} by
another (i.e., whether the latter is more general than the former),
while those for the assertive part establish whether its set of assertion
is \emph{consistent} (i.e., it has a model or it entails that a certain
individual is an instance of a given concept). Satisfiability tests
for descriptions and consistency tests for a set of assertion allow
one to establish whether the knowledge base is meaningful or not,
whereas subsumption tests allow one to maintain a hierarchy of concepts\emph{
ab universali}; finally, instance tests give one the ability of querying
the system against single individuals.

~

More formally, a generic DL axiom is a formula of one of the following
types:
\begin{itemize}
\item $C\equiv D$ (equivalence between concepts)
\item $C\sqsubseteq D$ (subsumption between concepts)
\item $R\equiv S$ (equivalence between roles)
\item $R\sqsubseteq S$ (subsumption between roles),%
\footnote{The equivalence and subsumption between roles may be indicated also
by the symbols $=$ and $\subseteq$, respectively.%
}
\end{itemize}
\noindent where the symbols $C,D$ are names or expressions of complex
concepts, which are formed by 1-ary or 2-ary operations on/between
atomic (indicated by $A$ in the following) or complex concepts, while
$R,S$ are names or expressions of complex roles which are formed
by 1-ary or 2-ary operations on/between atomic (indicated by $P$
in the following) or complex roles.

A generic DL assertion is a formula of one of the following types:
\begin{itemize}
\item $C\left(a\right)$ (concept assertion)
\item $R\left(a,b\right)$ (role assertion),
\end{itemize}
\noindent where the symbols $a,b$ are names of individuals, for which
the concept $C$ or the role $R$ holds.%
\footnote{A large number of constructs is listed in the table at the end of
this section.%
}

~

From a semantical point of view, an \emph{interpretation} \I is a
pair $(\deltaI,\cdot\ifunc)$, where the non-empty set \deltaI represents
the \strong{domain} of the interpretation and the \strong{interpretation function}
$\cdot\ifunc$ associates a set $\AI\subseteq\deltaI$ to every atomic
concept $A$ , a relation $\PI\subseteq\deltaII$ to every atomic
role $P$, and an element $\aI\in\deltaI$ to every individual $a$.
We write:%
\footnote{In the following, $\I\models\phi$ means ``$\I$ satisfies $\phi$,
where $\phi$ can be an axiom or an assertion. A syntax/semantical
reference for the main axioms, assertions and property declarations
is listed in the table at the end of this section.%
}
\begin{itemize}
\item $\I\models\left(C\equiv D\right)$ iff%
\footnote{``Iff'' is short for ``if and only if''.%
} $\CI=\DI$
\item $\I\models\left(C\sqsubseteq D\right)$ iff $\CI\subseteq\DI$
\item $\I\models\left(R\equiv S\right)$ iff $\RI=\SI$
\item $\I\models\left(R\sqsubseteq S\right)$ iff $\RI\subseteq\SI$
\item $\I\models C\left(a\right)$ iff $\aI\in\CI$
\item $\I\models R\left(a,b\right)$ iff $\abI\in\RI$
\end{itemize}
Finally, we say that \I is a \emph{model} for a DLKB \K (and write
$\I\models\K$) if \I satisfies all the axioms and assertions of
\K. The latter is said to be \emph{consistent} if there exists at
least an interpretation satisfying it, and the search of this interpretation
(\emph{consistency problem}) is just the clue to \emph{reasoning.}

\subsection{Families of logics}

\hspace*{0.7cm}As already observed, distinct DLs are characterized
by the constructs allowed to form complex concepts and roles starting
from atomic ones. The names are usually specified by a series of alphabet
letters and symbols. In the following, we will not concentrate on
formal semantics, but, for the sake of clarity, we will only hint
at the meaning of some constructs. 

By way of an example, we briefly describe the \alc logic (\emph{Attributive
Language with Complements}). Its syntax obeys the following rules:
\[
C,D\rightarrow A\ \vert\ \top\ \vert\ \bot\ \vert\ \neg C\ \vert\ C\sqcap D\ \vert\ C\sqcup D\ \vert\ \forall R.C\ \vert\ \exists R.C\,\,,
\]
where $\top$ denotes the concept enclosing any other one (\emph{top
concept}), $\bot$ denotes the concept enclosed in any other one (\emph{bottom
concept}), $\neg$ negates a concept; $\sqcup$ represents the union
of concepts (notice the analogy with the corresponding set operators)
and $\sqcap$ represents the intersection of concepts. The last two
constructs are called respectively \emph{universal }and \emph{existential
restriction}, and are pivotal in research connected to this report.
The concepts that can be built in \alc are called \alc-concepts.The
axioms\emph{ }and assertions that may be expressed in the \alc logic
are

\[
C\equiv D,\ \,\ \,\ C\sqsubseteq D,\ \,\ \,\ C\left(a\right),
\]
which indicate, respectively, equivalence and subsumption (also called
\emph{GCI}, \emph{General Concept Inclusion}) between two concepts,
and the membership of a concept. However, \emph{reasoning} in \alc
has a \noun{PSpace}-complete computational complexity.

In addition to those seen above, the most common constructs for concepts
which can be formed in DLs are $\le1.R$ (\emph{functional restriction},
denoted \Fu, that is equivalent to $\exists R.\top$), $\le nR$
and $\ge nR$ (\emph{numerical restrictions }, \Nu, which enclose
the functional one), $\le nR.C$ and $\ge nR.C$ (\emph{qualified
restrictions}, \Qu, that include the numerical ones), $\left\{ a\right\} $
and $\left\{ a_{1},\ldots,a_{n}\right\} $ (\emph{nominals},\Oo),
and $\exists R.\mathit{Self}$ (\emph{self-concept}). Contructs for
roles are often denoted by an operator symbol inside brackets after
the name of the logic. They are $R\supm$ (inverse role, \I), $R\cup S,R\cap S$
and $\neg R$ (role union, intersection and complement, respectively),
$R\circ S$ (role composition, also used in chain), $R\supp$ and
$R\sups$ (transitive and reflexive-transitive closure of roles),
$\mathit{id}\left(C\right)$(\emph{concept identity}), and $R_{C|\ensuremath{}},R_{|D}$
and $R_{C|D}$ (role restrictions). Sometimes, the symbols $U$ (or
\foreignlanguage{italian}{\emph{\small $\nabla$}},\emph{ universal
role}, defining the role which encloses any other one) and \emph{$N$
}(or \foreignlanguage{italian}{\emph{\strong{\emph{\footnotesize $\triangle$}}}},\emph{
empty role}, defining the role enclosed in any other one) are used.
Other kinds of axioms which can be introduced are $R\equiv S$ (equivalence
between roles), $R\sqsubseteq S$ (hierarchy between roles, \Hi,
denoted also by $R\subseteq S$), and the relative assertion $R\left(a,b\right)$
(role instance). The reflexive, irreflexive, symmetric, antisymmetric,
transitive, intransitive, disjunctive (two roles having no pair of
elements in common), functional and inverse functional properties
for roles are denoted by symbols that often differ in literature,
but are never ambiguous, e.g. respectively $\mathit{Sym}\left(R\right),\mathit{Asym}\left(R\right),\mathit{Refl}\left(R\right),\mathit{Irrefl}\left(R\right),\mathit{Tr}\left(R\right),\mathit{Intr}\left(R\right),\mathit{Disj}\left(R\right),\mathit{Fn}\left(R\right),\mathit{InvFn}\left(R\right)$.
When transitive property is allowed, we use the symbol \Sx, which
corresponds to $\alc(\supp)$. Sometimes, even small differences among
logics (e.g., limitation on the use of atomic rather than complex
roles on the right or left part of an assertion) can make a big difference
in expressiveness and complexity. Thus, it is not always easy to concisely
denote DLs, that consequently constitute an ever-open research field.

One or more among the above descripted constructs, axioms and assertions
are present in the families of logics that are the base of languages
used in ontologies. Among these, an important example is the logic
\sroiqd (in literature always denoted by \SxroiqD), at the bottom
of the \OWLtwoDL profile, which will be discussed in the following
(the letter in parenthesis indicates the use of \emph{concrete domains},
that will not be discussed here). Its syntax is easily inferred from
the symbols, while its complexity is \noun{N2ExpTime}-hard.%
\footnote{By the symbol \Ro, we mean that a DL allows complex inclusions of
the kind $R\circ S\sqsubseteq R$ and $R\circ S\sqsubseteq S$.%
}

~

\selectlanguage{italian}%
\noindent \begin{center}
\begin{tabular}{|c|c|c|}
\hline 
\emph{\footnotesize Construct} & \emph{\footnotesize Syntax} & \emph{\footnotesize Semantics}\tabularnewline
\hline 
\hline 
{\scriptsize top concept} & {\scriptsize $\top$} & {\scriptsize $\top\ifunc=\deltaI$}\tabularnewline
\hline 
{\scriptsize bottom concept} & \multirow{1}{*}{{\scriptsize $\bot$}} & {\scriptsize $\bot\ifunc=\emptyset$}\tabularnewline
\hline 
{\scriptsize concept negation} & {\scriptsize $\neg C$} & {\scriptsize $\left(\neg C\right)\ifunc=\deltaI\backslash\CI$}\tabularnewline
\hline 
{\scriptsize concept intersection} & {\scriptsize $C\sqcap D$} & {\scriptsize $\left(C\sqcap D\right)\ifunc=\CI\cap\DI$}\tabularnewline
\hline 
{\scriptsize concept union} & {\scriptsize $C\sqcup D$} & {\scriptsize $\left(C\sqcup D\right)\ifunc=\CI\cup\DI$}\tabularnewline
\hline 
{\scriptsize universal restriction} & {\scriptsize $\forall R.C$} & {\scriptsize $\left(\forall R.C\right)\ifunc=\left\{ a\in\deltaI\vert\left(\forall\left(a,b\right)\in\RI\right)\left(b\in C\ifunc\right)\right\} $}\tabularnewline
\hline 
{\scriptsize existential restriction} & {\scriptsize $\exists R.C$} & {\scriptsize $\left(\exists R.C\right)\ifunc=\left\{ a\in\deltaI\vert\,\exists\left(a,b\right)\in\RI\wedge b\in\CI\right\} $}\tabularnewline
\hline 
{\scriptsize numerical restriction} & {\scriptsize }%
\begin{tabular}{c}
{\scriptsize $\le nR$}\tabularnewline
{\scriptsize $\ge nR$}\tabularnewline
\end{tabular} & {\scriptsize }%
\begin{tabular}{c}
{\scriptsize $\left(\le nR\right)\ifunc=\left\{ a\in\deltaI\vert\,\#\left\{ b\in\deltaI|\left(a,b\right)\in\RI\right\} \le n\right\} $}\tabularnewline
{\scriptsize $\left(\ge nR\right)\ifunc=\left\{ a\in\deltaI\vert\,\#\left\{ b\in\deltaI|\left(a,b\right)\in\RI\}\right\} \ge n\right\} $}\tabularnewline
\end{tabular}\tabularnewline
\hline 
{\scriptsize qualified restriction} & {\scriptsize }%
\begin{tabular}{c}
{\scriptsize $\le nR.C$}\tabularnewline
{\scriptsize $\ge nR$.C}\tabularnewline
\end{tabular} & {\scriptsize }%
\begin{tabular}{c}
{\scriptsize $\left(\le nR.C\right)\ifunc=\left\{ a\in\deltaI\vert\,\#\left\{ b\in\CI|\left(a,b\right)\in\RI\right\} \le n\right\} $}\tabularnewline
{\scriptsize $\left(\ge nR.C\right)\ifunc=\left\{ a\in\deltaI\vert\,\#\left\{ b\in\CI|\left(a,b\right)\in\RI\right\} \ge n\right\} $}\tabularnewline
\end{tabular}\tabularnewline
\hline 
{\scriptsize nominals} & {\scriptsize }%
\begin{tabular}{c}
{\scriptsize $\left\{ a\right\} $}\tabularnewline
{\scriptsize $\left\{ a_{1},\ldots,a_{n}\right\} $}\tabularnewline
\end{tabular} & {\scriptsize }%
\begin{tabular}{c}
{\scriptsize $\left\{ a\right\} \ifunc=\left\{ \aI\right\} $}\tabularnewline
{\scriptsize $\left\{ a_{1},\ldots,a_{n}\right\} \ifunc=\left\{ a_{1}\ifunc,\ldots,a_{n}\ifunc\right\} $}\tabularnewline
\end{tabular}\tabularnewline
\hline 
{\scriptsize self concept} & {\scriptsize $\exists R.\mathit{Self}$} & {\scriptsize $\left(\exists R.Self\right)\ifunc=\left\{ a\in\deltaI\vert\left(a,a\right)\in\RI\right\} $}\tabularnewline
\hline 
{\scriptsize universal role} & {\scriptsize $U$, }{\small $\nabla$} & {\scriptsize $U\ifunc=\deltaII$}\tabularnewline
\hline 
{\scriptsize empty role} & {\scriptsize $N$,} \strong{{\footnotesize $\triangle$}} & {\scriptsize $N\ifunc=\emptyset\times\emptyset$}\tabularnewline
\hline 
{\scriptsize role inverse} & {\scriptsize $R\supm$} & {\scriptsize $\left(R\supm\right)\ifunc=\left\{ \left(a,b\right)\vert\left(b,a\right)\in\RI\right\} $}\tabularnewline
\hline 
{\scriptsize role negation} & {\scriptsize $\neg R$} & {\scriptsize $\left(\neg R\right)\ifunc=\left(\deltaII\right)\backslash\RI$}\tabularnewline
\hline 
{\scriptsize role intersection} & {\scriptsize $R\sqcap S$} & {\scriptsize $\left(R\sqcap S\right)\ifunc=\RI\cap\SI$}\tabularnewline
\hline 
{\scriptsize role union} & {\scriptsize $R\sqcup S$} & {\scriptsize $\left(R\sqcup S\right)\ifunc=\RI\cup\SI$}\tabularnewline
\hline 
{\scriptsize chaining} & {\scriptsize $R\circ S$} & {\scriptsize $\left(R\circ S\right)\ifunc=\RI\circ\SI$}\tabularnewline
\hline 
{\scriptsize concept identity} & {\scriptsize $\mathit{id}\left(C\right)$} & {\scriptsize $\left(id\left(C\right)\right)\ifunc=\left\{ \left(a,a\right)\vert\, a\in\CI\right\} $}\tabularnewline
\hline 
{\scriptsize transitive closure} & {\scriptsize $R\supp$} & {\scriptsize $\left(R\supp\right)\ifunc=\left(\RI\right)\supp$}\tabularnewline
\hline 
{\scriptsize refl-trans closure} & {\scriptsize $R\sups$} & {\scriptsize $\left(R\sups\right)\ifunc=\left(R\supp\right)\ifunc\cup\left(id\left(\top\right)\right)\ifunc$}\tabularnewline
\hline 
{\scriptsize role restriction} & {\scriptsize }%
\begin{tabular}{c}
{\scriptsize $R_{C|}$}\tabularnewline
{\scriptsize $R_{|D}$}\tabularnewline
{\scriptsize $R_{C|D}$}\tabularnewline
\end{tabular} & {\scriptsize }%
\begin{tabular}{c}
{\scriptsize $(R_{C|})\ifunc=\left\{ \left(a,b\right)\in\RI\vert\, a\in\CI\right\} $}\tabularnewline
{\scriptsize $(R_{|D})\ifunc=\left\{ \left(a,b\right)\in\RI\vert\, b\in\DI\right\} $}\tabularnewline
{\scriptsize $(R_{C|D})\ifunc=\left\{ \left(a,b\right)\in\RI\vert\, a\in\CI\wedge b\in\DI\right\} $}\tabularnewline
\end{tabular}\tabularnewline
\hline 
\end{tabular}
\par\end{center}

\noindent \begin{center}
\emph{\footnotesize Table 1.1.}{\footnotesize{} Main constructs for
concepts and}\textcolor{blue}{\footnotesize{} }{\footnotesize roles}
\par\end{center}{\footnotesize \par}

\noindent \begin{center}
\begin{tabular}{|c|c|}
\hline 
\emph{\footnotesize Syntax} & \emph{\footnotesize Semantics}\tabularnewline
\hline 
\hline 
{\small $C\equiv D$} & {\small $\CI=\DI$}\tabularnewline
\hline 
{\small $C\sqsubseteq D$} & {\small $\CI\subseteq\DI$}\tabularnewline
\hline 
{\small }%
\begin{tabular}{c}
{\small $C\left(a\right)$}\tabularnewline
{\small $\neg C\left(a\right)$}\tabularnewline
\end{tabular} & {\small }%
\begin{tabular}{c}
{\small $\aI\in\CI$}\tabularnewline
{\small $\aI\notin\CI$}\tabularnewline
\end{tabular}\tabularnewline
\hline 
{\small $R\equiv S$} & {\small $\RI=\SI$}\tabularnewline
\hline 
{\small $R\sqsubseteq S$} & {\small $\RI\subseteq\SI$}\tabularnewline
\hline 
{\small }%
\begin{tabular}{c}
{\small $R\left(a,b\right)$}\tabularnewline
{\small $\neg R\left(a,b\right)$}\tabularnewline
\end{tabular} & {\small }%
\begin{tabular}{c}
{\small $\abI\in\RI$}\tabularnewline
{\small $\abI\notin\RI$}\tabularnewline
\end{tabular}\tabularnewline
\hline 
{\small $R\sqcap S=N$, ~$\mathit{Disj}\left(R,S\right)$} & {\small $\RI\cap\SI=\emptyset\times\emptyset$}\tabularnewline
\hline 
{\small }%
\begin{tabular}{c}
{\small $\mathit{Refl}\left(R\right)$}\tabularnewline
{\small $\neg\mathit{Refl}\left(R\right)$}\tabularnewline
\end{tabular} & {\small }%
\begin{tabular}{c}
{\small $\left(\forall a\in\deltaI\right)\left(\left(a,a\right)\in\RI\right)$}\tabularnewline
{\small $\left(\forall a\in\deltaI\right)\left(\left(a,a\right)\notin\RI\right)$}\tabularnewline
\end{tabular}\tabularnewline
\hline 
{\small }%
\begin{tabular}{c}
{\small $\mathit{Sym}\left(R\right)$}\tabularnewline
{\small $\neg\mathit{Sym}\left(R\right)$}\tabularnewline
\end{tabular} & {\small }%
\begin{tabular}{c}
{\small $\left(\forall\left(a,b\right)\in\RI\right)\left(\left(b,a\right)\in\RI\right)$}\tabularnewline
{\small $\left(\forall\left(a,b\right)\in\RI\right)\left(\left(b,a\right)\notin\RI\right)$}\tabularnewline
\end{tabular}\tabularnewline
\hline 
{\small }%
\begin{tabular}{c}
{\small $\mathit{Trans}\left(R\right)$}\tabularnewline
{\small $\neg\mathit{Trans}\left(R\right)$}\tabularnewline
\end{tabular} & {\small }%
\begin{tabular}{c}
{\small $\left(\forall\left(a,b\right)\in\RI\right)\left(\left(b,c\right)\in\RI\rightarrow\left(a,c\right)\in\RI\right)$}\tabularnewline
{\small $\left(\forall\left(a,b\right)\in\RI\right)\left(\left(b,c\right)\in\RI\rightarrow\left(a,c\right)\notin\RI\right)$}\tabularnewline
\end{tabular}\tabularnewline
\hline 
\end{tabular}
\par\end{center}

\noindent \begin{center}
\emph{\footnotesize Table 1.2.}{\footnotesize{} Main types of axioms
and assertions on concepts and roles}\textcolor{blue}{\footnotesize }%
\footnote{The use of $\mathit{Trans}\left(R\right)$, $\mathit{Refl}\left(R\right)$,
and so on, looks like adding a new type of axiom. Actually, they must
be regarded as abbreviations of more complex expressions involving
equivalence, subsumption and various constructs which we will not
be treated here.%
}
\par\end{center}{\footnotesize \par}

\selectlanguage{british}%

\subsection{Semantic Web}

\hspace*{0.7cm}The so-called Web 2.0 describes the current model
of information search. The difference with Web 1.0 is represented
by the change in the origin of this information (from below\textminus{}users\textminus{}rather
than from above\textminus{}webmasters). Yet, a paradigm change in
its fundamental structure has not occured, as the creator of Web,
Tim Berners-Lee, wished instead. Indeed, contents are still exclusively
made up of pages connected by links, whose nontrivial words, together
with metadata, are indexed by search engines, by means of which a
correspondence between those and the one inserted by users can be
found. In this way, notwithstanding the high efficiency and precision
of the algorithms involved, such engines constitute a ``stupid''
example of \emph{information} \emph{retrieval}, where the matching
between terms, even if advanced, weighs more than their meaning.

The change brought by the so-called Semantic Web (\emph{SW} or Web
3.0) addresses the main issue that in the Web, as we know it, most
of the contents are structured in order to be read by humans, rather
than investigated by programs in an automatic way. A computer can
jump from page to page by following links, but it does not make any
assumption on the semantics of their contents. SW is an extension
of ``previous'' Web that allows one to assign a meaning to information
and provides machines with the ability to elaborate and \textquotedblleft{}understand\textquotedblright{}
what in the past they could only show. For all this to work, computers
should be able to access well-structured information schemas and apply
inference rules to permit automatic reasoning, in order to delocalize
knowledge representation and spread it over the Net. The languages
these rules apply should be expressive enough to make the Web capable
of \textquotedblleft{}reasoning\textquotedblright{} in a versatile
and widespread way. That is made possibile by tools such as \emph{eXstensible
Markup Language} (\emph{XML}), \emph{Resource Description Framework}
(\emph{RDF}) and the ontologies described by the \emph{Ontology Web
Language} (\emph{OWL}).

\subsection{RDF graphs}

\hspace*{0.7cm}RDF is a model for representation of information on
the Web. The basic idea is that every resource (concept, class, property,
object, value, etc.) can be univocally described in terms of simple
or identified-by-value properties. All this permits to schematize
an RDF model by means of an oriented graph, whose nodes represent
primitive objects and whose edges denote properties. The limited vocabulary
of RDF is extended by \emph{RDFS} (\emph{RDF Schema}), which allows
one to define classes and properties in a more powerful way. E.g.,
it is possibile to define a property as a relation by indicating its
domain and/or range, or say what class is a subclass of another.

From a data structure point of view, an RDF graph may be seen as a
set of triples of type \code{(subject predicate object)}. For each
triple,\code{ subject }and\code{ object }are two nodes connected
by an edge that represents a predicate; thus, the subject of a triple
may be the object of another and viceversa. Each of these nodes may
be identified by a URI/IRI, and so it can represent a resource, or
it may be simply a ``connector'' (in this case, it is called a \emph{blank
node}, \emph{bnode}, or \emph{anonymous node}). If an object is not
a subject of another triple, it may also be a \emph{literal}, i.e.,
a datum representing a value in a certain domain.

On the representation side, W3C suggests five types of \emph{serialization}s
for an RDF graph, the most used of which are~ \textit{RDF/XML} (XML
language is used) and \textit{Turtle} (descriptions through lists
of triples).%
\footnote{The \emph{serialization} of an object is the process which allows
one to represent it in an accessible way, in our case a text file.%
} XML is used because of its precise representation of data. Thus,
they can be shared, by means of RDF, among various Web applications
while preserving their original meaning.

\subsection{Ontologies}

\hspace*{0.7cm}OWL ontologies extend RDFS vocabulary further.%
\footnote{Computer science ambitiously draw on the philosophical term \emph{ontology}
to show that the OWL language can describe the world starting from
its ultimate constituents and from relations among them.%
} If RDFS guarantees generality and precision in constructing knowledge
representation, with OWL we reach such an expressiveness that we can
``argue'' about described information and ``extract'' more, through
\emph{reasoning}. It provides powerful tools to define classes (that
represent concepts), properties (relations among classes) and individuals
(their instances), and gives the chance to combine them in logic constraints
by means of which necessity and/or sufficiency may be expressed. 

Upon a W3C \emph{recommendation}, OWL is made up of three sublanguages
with increasing expressiveness: \OWLLite, \OWLDL and \OWLFull.
The first two have an almost total correspondence between two peculiar
DLs (resp.\emph{ \shifd }and\emph{ \shoind}).%
\footnote{OWL-Lite and OWL-DL provide the possibility of defining \emph{annotation}s,
which correspondent DLs do not do. In any case, annotations affect
neither \emph{reasoning}, nor complexity, nor decidability. For the
sake of precision, OWL-Lite is considered a ``syntactic subset''
of OWL-DL.%
} The last one uses the same subset of \OWLDL constructs, but allows
them to be unconstrained, according to RDF style. Due to the lack
of restrictions on transitive property and to the possibility of handling
concepts as individuals (\emph{metamodeling}), \OWLFull turns out
to be undecidable and thus, for the correspondence between that and
any of the aforesaid DLs, \OWLDL is the most expressive decidable
sublanguage of OWL.%
\footnote{Obviously, nothing excludes there is another decidable and more expressive\textminus{}but
not studied yet\textminus{}sublanguage.%
} 

OWL 2 extends OWL by enhancing its features and expressiveness. Apart
from a limited number of cases, OWL 2 is perfectly ``backward compatible''
with the old OWL (which we call OWL 1 to differentiate it from the
OWL 2), i.e., all the ontologies of the latter keep the same semantics
as that they had before, even if ``fed'' to a \emph{reasoner} for
OWL 2. Concisely, OWL 2 introduces a simplification into the writing
of the most common \emph{statements}, it permits \emph{metamodeling},
together with some new constructs which increase its expressiveness,
and it extends support to data types.

As for sublanguages, in OWL 2 we can deal with \OWLtwoFull and \OWLtwoDL.
As mentioned before, the latter corresponds to the \emph{\sroiqd}
logic. Though the former may be syntactically seen as the union of
the latter with RDFS, it is semantically compatible with \OWLtwoDL
(insofar as its semantics allows one to draw all the inferences that
can be drawn by using the semantics of \OWLtwoDL). Similarly to the
analogue sublanguage \OWLoneFull, OWL 2 finds application in the
modelling of concepts where \emph{reasoning} is not required. A new
feature of OWL 2 is the use of \emph{profiles} (i.e., fragments of
language) \OWLtwoEL, \OWLtwoQL and \OWLtwoRL:%
\footnote{In this way, \OWLoneLite, \OWLoneDL and \OWLtwoDL may be considered
profiles of OWL 2.%
} the first is useful for applications involving ontologies that contain
a great number of classes and/or properties; the second aims at those
applications with large volumes of instances, where answering the
queries is of primary importance (from which the name QL, \emph{Query
Language}); the last profile is useful in applications not requiring
the sacrifice of too much expressiveness to efficiency. We will not
delve into the structural characteristics of these profiles. Here,
we only say that they are grounded on very different DLs, each of
them fit for the specific purpose they are designed for.

\subsection{Syntactical correspondence of OWL(2) with DLs}

\hspace*{0.7cm}As to its practical representation, an ontology is
a RDF/XML-structured text file, but, for our goals, it will be more
useful to think to it in terms of an RDF graph and, thus, of triples.
Table 1.3 describes the correspondence between the most common constructs
of DLs and such triples (or a single resource, where applicable),
which use RDF(S) syntax together with that of OWL(2). It also contains
the names given by OWL to the fundamental concepts and roles of DLs.
It is interesting to notice how a DL construct often corresponds to
more triples, which sometimes makes interpretation quite hard. The
symbols \code{\_:x} and \code{\_list} in the Table 1.3 indicate
respectively a \emph{blank node} and a \emph{list} (a structure we
will not deal with), whereas the prefixes before the elements of triples
represent standard W3C \emph{namespaces}, which serve as abbreviations
for the URIs they refer to.%
\footnote{The correspondences between namespaces and prefixes are: 
\begin{elabeling}{00.00.0000}
\item [{\hspace*{0.7cm}\code{rdf:~~~http://www.w3.org/1999/02/22-rdf-syntax-ns\#}}]~
\item [{\hspace*{0.7cm}\code{rdfs:~~http://www.w3.org/2000/01/rdf-schema\#}}]~
\item [{\hspace*{0.7cm}\code{owl:~~~http://www.w3.org/2002/07/owl\#}}]~
\item [{\hspace*{0.7cm}\code{xsd:~~~http://www.w3.org/2001/XMLSchema\#}}]~\end{elabeling}
}

\subsection{SPARQL}

\hspace*{0.7cm}One of the most widespread languages for querying
against ontologies is \emph{SPARQL} \emph{(}recursive acronym for
\emph{SPARQL Protocol And RDF Query Language}). Although it has lots
of analogies with SQL, most of which syntactical, it is equipped with
a fundamentally different semantics. As query languages for databases
essentially work on tables and handle logic conditions to select the
rows of these tables that satisfy them, the\code{ WHERE }clause in
SPARQL finds matches between the triples of the query and those of
the ontology indicated in the\code{ FROM }clause. The logic assessments
are relegated to the\code{ FILTER }operator, which possesses a big
expressive power, thanks to its several functions, but is rarely used
due to the loss of efficiency that its presence may cause.%
\footnote{Depending on implementation, the\code{ FROM }clause may be implied
because software loads in memory the ontology model separately.%
} The\code{ SELECT }clause is very similar to that of SQL. It accepts
\code{ DISTINCT }as a keyword and the classical aggregation operators
(\code{COUNT},\code{ SUM},\code{ MIN},\code{ AVG } and\code{ MAX}),
while the\code{ GROUP BY }and\code{ HAVING }clauses are often used
(another similarity) to refine the selection. The involved variables
are preceded by the symbol '\code{?}', whereas constants do not have
a particular syntax, even if they generally coincide with URI/IRIs
of resources present in the ontology one is handling.

Analyzing in finer detail the\code{ WHERE }clause, one can say that
the triples to match\emph{ }are enclosed in a block between braces
and separated by a dot, which implies their intersection. Inside these
braces, more sub-blocks may be found, which are useful in making the
union (\code{UNION }operator) and the difference (\code{MINUS }operator)
between sets. The matching is valid if the variables having the same
names in the clause have the same values in the ontology. There exist
some useful shortenings, such as the use of semicolon or colon in
place of dot, which act respectively in the following way:

\begin{center}
\code{?s ?p1 ?o1 ; ?p2 ?o2 . }shortens\code{ ?s ?p1 ?o1 . ?s ?p2 ?o2 .}
\par\end{center}

\begin{center}
\code{?s ?p ?o1 , ?o2 . }shortens\code{ ?s ?p ?o1 . ?s ?p ?o2 .}
\par\end{center}

In addition to selection, one can also have the\code{ DESCRIBE},\code{ ASK }
and\code{ CONSTRUCT }query types, which we will not review here.
The employment of the\code{ PREFIX }clauses, that precede the real
query and indicate the abbreviations for namespaces used inside of
it, is quite peculiar.

\begin{center}
\begin{tabular}{|c|c|}
\hline 
\emph{\small DL construct} & \emph{\small RDF(S)/OWL(2) resource/triple(s)}\tabularnewline
\hline 
\hline 
\texttt{\small $\top$} & \texttt{\small owl:Thing}\tabularnewline
\hline 
\texttt{\small $\bot$} & \texttt{\small owl:Nothing}\tabularnewline
\hline 
\texttt{\small $C\sqsubseteq D$} & \texttt{\small ($C$ rdfs:subClassOf $D$)}\tabularnewline
\hline 
\texttt{\small $C\equiv D$} & \texttt{\small ($C$ owl:equivalentClass $D$)}\tabularnewline
\hline 
\texttt{\small $C\sqsubseteq\neg D$}\texttt{\footnotesize{} ~}{\footnotesize opp.}\texttt{\small{} $C\sqcap D\sqsubseteq\bot$} & \texttt{\small ($C$ owl:disjointWith $D$)}\tabularnewline
\hline 
\texttt{\small $C_{1}\sqcap C_{2}\sqcap\ldots\sqcap C_{n}$} & \texttt{\small (\_:x owl:intersectionOf \_list$\left(C_{1},C_{2},\ldots,C_{n}\right)$)}\tabularnewline
\hline 
\texttt{\small $C_{1}\sqcup C_{2}\sqcup\ldots\sqcup C_{n}$} & \texttt{\small (\_:x owl:unionOf \_list$\left(C_{1},C_{2},\ldots,C_{n}\right)$)}\tabularnewline
\hline 
\texttt{\small $\neg C$} & \texttt{\small (\_:x owl:complementOf $C$)}\tabularnewline
\hline 
\texttt{\small $\left\{ a_{1},a_{2},\ldots,a_{n}\right\} $} & \texttt{\small (\_:x owl:oneOf \_list$\left(a_{1},a_{2},\ldots,a_{n}\right)$)}\tabularnewline
\hline 
\texttt{\small $\exists R.C$} & \texttt{\small }%
\begin{tabular}{c}
\texttt{\small (\_:x owl:someValuesFrom $C$)~(\_:x owl:onProperty
$R$)}\tabularnewline
\end{tabular}\tabularnewline
\hline 
\texttt{\small $\forall R.C$} & \texttt{\small }%
\begin{tabular}{c}
\texttt{\small (\_:x owl:allValuesFrom $C$)~(\_:x owl:onProperty
$R$)}\tabularnewline
\end{tabular}\tabularnewline
\hline 
\texttt{\small $\exists R.\left\{ a\right\} $} & \texttt{\small }%
\begin{tabular}{c}
\texttt{\small (\_:x owl:hasValue $a$)~(\_:x owl:onProperty $R$)}\tabularnewline
\end{tabular}\tabularnewline
\hline 
\texttt{\small $\le nR$} & \texttt{\small }%
\begin{tabular}{c}
\texttt{\small (\_:x owl:minCardinality $n$)~(\_:x owl:onProperty
$R$)}\tabularnewline
\end{tabular}\tabularnewline
\hline 
\texttt{\small $\ge nR$} & \texttt{\small }%
\begin{tabular}{c}
\texttt{\small (\_:x owl:maxCardinality $n$)~(\_:x owl:onProperty
$R$)}\tabularnewline
\end{tabular}\tabularnewline
\hline 
\texttt{\small $\le nR\,\sqcap\ge nR$} & \texttt{\small }%
\begin{tabular}{c}
\texttt{\small (\_:x owl:cardinality $n$)~(\_:x owl:onProperty $R$)}\tabularnewline
\end{tabular}\tabularnewline
\hline 
\texttt{\small $\le nR.C$} & \texttt{\small }%
\begin{tabular}{c}
\texttt{\small (\_:x owl:minQualifiedCardinality $n$)}\tabularnewline
\texttt{\small (\_:x owl:onClass $C$)~(\_:x owl:onProperty $R$)}\tabularnewline
\end{tabular}\tabularnewline
\hline 
\texttt{\small $\ge nR.C$} & \texttt{\small }%
\begin{tabular}{c}
\texttt{\small (\_:x owl:maxQualifiedCardinality $n$)}\tabularnewline
\texttt{\small (\_:x owl:onClass $C$)~(\_:x owl:onProperty $R$)}\tabularnewline
\end{tabular}\tabularnewline
\hline 
\texttt{\small $\le nR.C\,\sqcap\ge nR.C$} & \texttt{\small }%
\begin{tabular}{c}
\texttt{\small (\_:x owl:qualifiedCardinality $n$)}\tabularnewline
\texttt{\small (\_:x owl:onClass $C$)~(\_:x owl:onProperty $R$)}\tabularnewline
\end{tabular}\tabularnewline
\hline 
\texttt{\small $\exists R.Self$} & \texttt{\small }%
\begin{tabular}{c}
\texttt{\small (\_:x owl:hasSelf }\texttt{\emph{\small true}}\texttt{\small )~(\_:x
owl:onProperty $R$)}\tabularnewline
\end{tabular}\tabularnewline
\hline 
\texttt{\small $C\left(a\right)$} & \texttt{\small ($a$ rdf:type $C$)}\tabularnewline
\hline 
\texttt{\small $\left\{ a_{i}\right\} \equiv\left\{ a_{j}\right\} $} & \texttt{\small ($a_{i}$ owl:sameAs $a_{j}$)}\tabularnewline
\hline 
\texttt{\small $\left\{ a_{i}\right\} \sqsubseteq\neg\left\{ a_{j}\right\} $} & \texttt{\small ($a_{i}$ owl:differentFrom $a_{j}$)}\tabularnewline
\hline 
\texttt{\small $U$ }\texttt{\footnotesize ~opp.}\texttt{\small{} }\texttt{\textifsymbol[ifgeo]{51}} & \texttt{\small owl:topObjectProperty}\tabularnewline
\hline 
\texttt{\small $N$ }\texttt{\footnotesize ~opp.}\texttt{\small{} }\texttt{\textifsymbol[ifgeo]{49}} & \texttt{\small owl:bottomObjectProperty}\tabularnewline
\hline 
\texttt{\small $R\sqsubseteq S$} & \texttt{\small ($R$ owl:subPropertyOf $S$)}\tabularnewline
\hline 
\texttt{\small $R\equiv S$} & \texttt{\small ($R$ owl:equivalentProperty $S$)}\tabularnewline
\hline 
\texttt{\small $R\sqcap S\sqsubseteq N$} & \texttt{\small ($R$ owl:PropertyDisjointWith $S$)}\tabularnewline
\hline 
\texttt{\small $R\supm$} & \texttt{\small (\_:x owl:inverseOf $R$)}\tabularnewline
\hline 
\texttt{\small $R_{1}\circ R_{2}\circ\ldots\circ R_{n}\sqsubseteq R$} & \texttt{\small ($R$ owl:propertyChainAxiom \_list$\left(R_{1},R_{2},\ldots,R_{n}\right)$)}\tabularnewline
\hline 
\texttt{\small $\mathit{Refl}\left(R\right)$ } & \texttt{\small ($R$ rdf:type owl:ReflexiveProperty)}\tabularnewline
\hline 
\texttt{\small $\mathit{Irrefl}\left(R\right)$} & \texttt{\small ($R$ rdf:type owl:IrreflexiveProperty)}\tabularnewline
\hline 
\texttt{\small $\mathit{Sym}\left(R\right)$} & \texttt{\small ($R$ rdf:type owl:SymmetricProperty)}\tabularnewline
\hline 
\texttt{\small $\mathit{Asym}\left(R\right)$} & \texttt{\small ($R$ rdf:type owl:AsymmetricProperty)}\tabularnewline
\hline 
\texttt{\small $\mathit{Trans}\left(R\right)$} & \texttt{\small ($R$ rdf:type owl:TransitiveProperty)}\tabularnewline
\hline 
\texttt{\small $\mathit{Fn}\left(R\right)$} & \texttt{\small ($R$ rdf:type owl:FunctionalProperty)}\tabularnewline
\hline 
\texttt{\small $\mathit{InvFn}\left(R\right)$} & \texttt{\small ($R$ rdf:type owl:InverseFunctionalProperty)}\tabularnewline
\hline 
\texttt{\small $R\left(a,b\right)$} & ($a$\code{\emph{ property\_name} }$b$)\tabularnewline
\hline 
\texttt{\small $\neg R\left(a,b\right)$} & \texttt{\small }%
\begin{tabular}{c}
\texttt{\small (\_:x rdf:type owl:NegativePropertyAssertion)}\tabularnewline
\texttt{\small (\_:x owl:sourceIndividual $a$)}\tabularnewline
\texttt{\small (\_:x owl:assertionProperty $R$)}\tabularnewline
\texttt{\small (\_:x owl:targetIndividual $b$)}\tabularnewline
\end{tabular}\tabularnewline
\hline 
\end{tabular}
\par\end{center}

\selectlanguage{italian}%
\begin{center}
\emph{\footnotesize Table 1.3. }{\footnotesize Correspondences between
DLs and SW}
\par\end{center}{\footnotesize \par}

\pagebreak{}

\selectlanguage{british}%

\section{\noindent \label{sec:ANALYSIS-AND-RESULTS}ANALYSIS AND RESULTS}

\subsection{The family \dl<\allpiz>}

\selectlanguage{italian}%
\thispagestyle{empty}\foreignlanguage{british}{\hspace*{0.7cm}Description
Logics derived from decidable fragments of set theory, generally denoted
by the notation}\linebreak{}
\foreignlanguage{british}{\dl<\noun{language\_name}>, are having
considerable importance. }

\selectlanguage{british}%
The family of logics underlying this class of ontologies, object of
research this project is based on, is focused on the fragment \allpiz~,
which has a good expressiveness w.r.t. knowledge representation in
real-world applications. The interest aroused by that is related to
NP-completeness of its decision procedure in some cases of practical
relevance. \dl<\allpiz>-concepts and \dl<\allpiz>-roles are formed
according to the syntax%
\footnote{The symbol $\text{sym}\left(R\right)$ denotes symmetric closure of
$R$.%
}
\begin{eqnarray*}
C,D & \rightarrow & A\ \vert\ \top\ \vert\ \bot\ \vert\ \neg C\ \vert\ C\sqcap D\ \vert\ C\sqcup D\ \vert\ \left\{ a\right\} \ \vert\ \exists R.Self\ \vert\ \exists R.\left\{ a\right\} \\
R,S & \rightarrow & P\ \vert\ U\ \vert\ R\supm\ \vert\ \neg R\ \vert\ R\sqcap S\ \vert\ R\sqcup S\ \vert\ R_{C|\ensuremath{}}\ \vert\ R_{|D}\ \vert\ R_{C|D}\ \vert\ id\left(C\right)\ \vert\ \text{{\text{sym}}}\left(R\right)\,,
\end{eqnarray*}
while a \dl<\allpiz>-KB is made of axioms and assertions of the
following kind:
\[
\begin{array}{c}
C\equiv D\ \,\ \,\ C\sqsubseteq D\ \,\ \,\ C\sqsubseteq\forall R.D\ \,\ \,\ \exists R.C\sqsubseteq D\ \,\ \,\ C\left(a\right)\ \,\ \,\ R\left(a,b\right)\\
R\equiv S\ \,\ \,\ R\sqsubseteq S\ \,\ \,\ R\circ R'\sqsubseteq S\ \,\ \,\ Trans\left(R\right)\ \,\ \,\ Refl\left(R\right)\ \,\ \,\ Asym\left(R\right)\ \,\ \,\ Fn\left(R\right)\ \,\ \,\ InvFn\left(R\right)\,.
\end{array}
\]
We may notice that universal restriction is allowed only in the right
part of a subsumption, whereas existential restriction can appear
only in the left part. In addition, neither numerical, nor qualified
restrictions are allowed.

\subsection{Description of the experiment}

\hspace*{0.7cm}Our analysis is aimed at selecting ontologies corresponding
to the \dl<\allpiz> family, in order to use them as a base of study
for this family of logics in real-world applications. To do that,
it is necessary to query against as many as possible publicly available
ontologies in a quick and efficient way. The use of queries in SPARQL
conveniently lends itself to the purpose. The management is provided
by \strong{dOWLphin}, a Java library specifically created in order
to easily load the ontologies and prepare the queries. The underlying
library is \strong{Jena}, one of the most common collections of API
for the Semantic Web, that is very handy because it can directly deal
with triples. As a \emph{front-end} GUI, the program \strong{QuAny}
was used, which was born inside the experiment too and allows one
to query against local and remote ontologies, and save on disk the
corresponding results, for future verification.

\subsection{Results and future objectives}

\hspace*{0.7cm}The query was employed to test a significant number
of ontologies of\strong{ BioPortal}, the web portal of the National
Center for Biomedical Ontology. This choice was not random, but motivated
mainly by two reasons: the first concerns the large amount of ontologies
present on the portal, coming from repositories of biomedical resources
spread all over the world; the second is relative to the connection
that these ontologies have with the real world in general, and with
human life and medicine in particular, which are fields offering several
matters for reflection on how widely knowledge of so important themes
may be schematized and represented, and, most of all, on the role
\emph{reasoning} may have in automatically inferring new information. 

Around 30\% of the ontologies resulted member of the language \dl<\allpiz>,
which brings good hopes for reasoning on them, given that\textminus{}we
remind\textminus{}w.r.t. computational complexity we are in the NP-completeness
realm. Concerning the remaining 70\%, efficient algorithms for conversion
will have to be considered and semantic tests will have to be done,
as previously happened for the \dl<\mlsscart> language (cfr. \citet{C+10}). 

In Appendix \ref{sec:app-b}, a table shows which ontologies were
recognized as members of the language \dl<\allpiz>. 

~

\pagebreak{}
\selectlanguage{italian}%
\begin{quotation}
\thispagestyle{empty}\end{quotation}
\selectlanguage{british}%

\appendix
\pagebreak{}

\section{\noindent \label{sec:app-a}ANALYSIS OF THE SPARQL QUERY}
\selectlanguage{italian}%
\begin{quotation}
\thispagestyle{empty}
\end{quotation}
\hypertarget{app}{}

\selectlanguage{british}%
As we are going to see in the implementation, to write declaratively
what an algorithm could do through a list of well-constructed statements
is not an immediate operation to undertake. The final query is the
result of various logical lines of argument concerning the conversion
from \dl<\allpiz>-constructs into those of RDF(S)/OWL(2), opportunely
applied to some test ontologies. 

From Table 1.3, one infers that, for the ontology under examination
to be considered part of the family of logics seen before, the allowed
elements of RDF(S)/OWL(2) are the following:
\begin{itemize}
\item class or property definition (\texttt{rdf:type})
\item top concept (\texttt{owl:Thing})
\item bottom concept (\texttt{owl:Nothing})
\item concept negation (\texttt{owl:complementOf})
\item intersection and union of concepts (resp. \texttt{owl:intersectionOf}
and \texttt{owl:unionOf)}
\item singleton (\texttt{owl:oneOf})
\item self concept (\texttt{owl:hasSelf})
\item value restriction (\texttt{owl:hasValue})
\item universal role (\texttt{owl:topObjectProperty})
\item role inverse (\texttt{owl:inverseOf)}
\item intersection and union of roles (not present in OWL 2)%
\footnote{Obviously, we may ignore the constructs not present in OWL 2, since
they cannot be a cause of rejection of the ontology.%
}
\item role restrictions (not present in OWL 2)
\item identity concept (not present in OWL 2)
\item symmetric closure (it can be emulated by means of union and inverse
of roles)
\item equivalence axiom between concepts (\texttt{owl:equivalentClass})
\item subsumption axiom between concepts (\texttt{rdfs:subClassOf})
\item exiistential restriction (\texttt{owl:someValuesFrom}) only in the
left part of a subsumption 
\item universal restriction (\texttt{owl:allValuesFrom}) only in the right
part of a subsumption 
\item minimal numerical restriction (\texttt{owl:minCardinality}) with cardinality
not greater than\code{ 1}, only in the left part of a subsumption
\item declaration of concept membership (\texttt{rdf:type})
\item declaration of role membership
\item equivalece between roles (\texttt{owl:equivalentProperty})
\item inclusion between roles (\texttt{owl:subPropertyOf})
\item role chaining (\texttt{owl:PropertyChainAxiom})
\item transitive property (\texttt{owl:TransitiveProperty})
\item reflexive property (\texttt{owl:ReflexiveProperty})
\item asymmetric property (\texttt{owl:AsymmetricProperty})
\end{itemize}
~

We have to add to these all the ones that can be deduced from a combination
of them, such as
\begin{itemize}
\item disjunction between concepts (\texttt{owl:disjointWith}, \texttt{owl:members})
\item union of disjoint concepts (\texttt{owl:disjointUnionOf})
\item equivalence and non-equivalence between individuals (resp. \texttt{owl:sameAs
}and\texttt{ owl:differentFrom})
\item empty role (\texttt{owl:bottomObjectProperty})
\item role domain (\texttt{rdfs:domain}, some cases excluded)%
\footnote{\label{fn:dominio}Role domain corresponds to the DLs' axiom $\exists R.\top\sqsubseteq C$,
so it is necessary to verify that $C$ is satisfactorily expressed
by the language.%
}
\item role range (\texttt{rdfs:range})
\item disjunction of properties (\texttt{owl:propertyDisjointWith}, \texttt{owl:members})
\item irreflexive property (\texttt{owl:IrreflexiveProperty})
\item symmetric property (\texttt{owl:SymmetricProperty})
\item direct and inverse functional property (resp. \texttt{owl:FunctionalProperty},
\texttt{owl:InverseFunctionalProperty})
\item declaration of role non-membership (\texttt{owl:NegativePropertyAssertion})
\end{itemize}
~

Consequently, the constructs which cause the rejection of an ontology
are:
\begin{itemize}
\item existential restriction not in the left part of a subsumption 
\item universal restriction not in the right part of a subsumption
\item minimal numerical restriction with cardinality greater than\code{ 1 }or
not in the left part of a subsumption
\item qualified numerical restrictions (\texttt{owl:{[}max|min{]}QualifiedCardinality},
\texttt{owl:qualifiedCardinality})
\item maximum and exact unqualified numerical restrictions (resp. \texttt{owl:maxCardinality},
\texttt{owl:cardinality})
\item some cases of domain declaration%
\footnote{Cf. footnote \prettyref{fn:dominio}.%
}
\item datatype (\texttt{rdfs:Datatype}, \texttt{owl:DatatypeProperty})
\end{itemize}
~

\hspace*{0.7cm}In order to create the query, we must consider that
the language \dl<\allpiz> is very expressive and provides constructs
and axioms which range over several elements of RDF(S)/OWL(2). Thus,
to abbreviate the task of enumerating them, the criterion of selecting
the triples not related to the language was adopted. Consequently,
the result of the query will contain at least a triple if the examined
ontology is to be rejected. There are as many clauses as the elements
to consider, connected by the\code{ UNION }operator; thus, if at
least a match is found, a non empty result will be obtained, i.e.,
the analyzed ontology will not be a member of the language.

Now, let us analyze the correctness of the query in detail (the relative
row number is indicated in brackets, when necessary).
\begin{itemize}
\item The first clause (lines 7-19) includes the triples corresponding to
existential restrictions (line 8). Yet, not all of them contribute
to the rejection of the ontology, but only those which are not in
the left part of a subsumption. Thus, in line 10 we force their exclusion,
provided that any is found. If this is not the case, we find a match,
according to line 8; otherwise, we must control that neither of the
following cases occurs (lines 11-17):%
\footnote{\label{fn:Doppio MINUS}The second-level\code{ MINUS }operator could
be optimized by including the various triples contained in its clauses
after line 8; nonetheless, we must not forget that the \emph{matching}
of triples is a notoriusly not very much efficient operation: the
choice of double-nesting concurs to avoid that SPARQL engine performs
useless searches when no subsumption in line 10 is found.%
}

\begin{itemize}
\item the anonymous node corresponding to a restriction is in the right
part of any triple (line 12): that allows one to include all the axioms
with the restriction and the various operators (e.g., union, intersection,
complement, etc.) which work on lists and single classes in the right
part;
\item the above-mentioned node is in the left part of a triple corresponding
to an axiom (lines 13-16): that allows one to include single and multiple
equivalences and disjunctions.
\end{itemize}
\item The second clause (lines 19-33) includes the triples corresponding
to universal restrictions (line 20). When a triple of this kind is
found, in order for a\emph{ matching} to be valid, it is also necessary
to check that at least either of the following cases occurs (lines
22-31):

\begin{itemize}
\item the anonymous node corresponding to a restriction is in the right
part of a triple (line 23), but this triple is not a subsumption (line
24), nor that node is the domain of a property (line 25);
\item the above-mentioned node is in the left part of a triple (line 27),
but it is not part of the definition of the restriction itself (lines
28-30) (otherwise, an ever-false condition would be represented):
that is to control that the restriction is neither in the left part
of an axiom, nor in the left part of some class construct.
\end{itemize}
\item The third clause (lines 33-46) includes the triples corresponding
to an operator of minimal cardinality (line 34). Here we proceed (line
36 and lines 39-43) as in the first clause, with the difference (line
37) that the cardinality value must be not greater than\code{ 1 }too.%
\footnote{Footnote \prettyref{fn:Doppio MINUS} holds also for the nested\code{ MINUS}es. %
}
\item The remaining clauses (lines 47-51) check for \emph{datatype} (lines
47 and 48), qualified cardinality (line 49: the\texttt{\small{} onClass
}property is peculiar to this kind of restriction), exact and maximum
cardinality (lines 50 and 51) triples.
\item For the purpose of accelerating the execution, the use of the\code{ LIMIT }operator
(line 52) reduces the size of results to only one element, which is
enough to reject the ontology.
\end{itemize}
~

\begin{lstlisting}[basicstyle={\ttfamily},numbers=left,numberstyle={\small},showstringspaces=false,tabsize=4]
PREFIX rdf:  <http://www.w3.org/1999/02/22-rdf-syntax-ns#>
PREFIX rdfs: <http://www.w3.org/2000/01/rdf-schema#>
PREFIX owl:  <http://www.w3.org/2002/07/owl#>

SELECT *
WHERE {	
		{ 	
			?left owl:someValuesFrom ?class .
			MINUS {	
				?left rdfs:subClassOf ?right . 
				MINUS {	
					{ ?s ?p ?left . } UNION
					{ ?left owl:equivalentClass ?right . } UNION
					{ ?left owl:disjointWith ?right . } UNION
					{ ?left owl:members ?right . } UNION 				
					{ ?left owl:disjointUnionOf ?right . }
				} 
			}	
		} UNION {	
			?restr owl:allValuesFrom ?class .
			{ 
				{		
					?x ?prop ?restr .
					MINUS { ?x rdfs:subClassOf ?restr . }
					MINUS { ?x rdfs:domain ?restr . }
				} UNION {	
					?restr ?prop ?x .
					MINUS { ?restr rdf:type ?x . } 
					MINUS { ?restr owl:onProperty ?x . }
					MINUS { ?restr owl:allValuesFrom ?x . }
				}
			}		
		} UNION { 	
			?left owl:minCardinality ?num .
			MINUS {	
				?left rdfs:subClassOf ?right . 
				FILTER (?num<=1) . 
				MINUS {	
					{ ?s ?p ?left . } UNION
					{ ?left owl:equivalentClass ?right . } UNION
					{ ?left owl:disjointWith ?right . } UNION 		
					{ ?left owl:members ?right . }	UNION 
					{ ?left owl:disjointUnionOf ?right . }
				}
			}		
		} UNION	
		{ ?s ?p owl:DatatypeProperty . } UNION	
		{ ?s ?p rdfs:Datatype . } UNION	
		{ ?s owl:onClass ?o . } UNION	
		{ ?s owl:cardinality ?o . } UNION	
		{ ?s owl:maxCardinality ?o . } 
} LIMIT 1	
\end{lstlisting}

\selectlanguage{italian}%
\begin{quotation}
\pagebreak{}
\end{quotation}
\selectlanguage{british}%

\section{\noindent \label{sec:app-b}TEST AGAINST BIOPORTAL ONTOLOGIES}

\selectlanguage{italian}%
\thispagestyle{empty}

\selectlanguage{british}%
The SPARQL query was tested against the ontologies of BioPortal, available
at the link \url{http://rest.bioontology.org/bioportal/virtual/download/ID?apikey=KEY}.
The API\code{ KEY }is an identifier for the registered users. The
following table associates the\code{ ID }of the previous URL to the
symbolic name of the corrispondent ontology and highlights the ones
which belong to the language.{\footnotesize }%
\begin{longtable}{|c|c|c||c|c|c||c|c|c|}
\hline 
\emph{\scriptsize ID} & \emph{\scriptsize Name} & \emph{\scriptsize Res} & \emph{\scriptsize ID} & \emph{\scriptsize Name} & \emph{\scriptsize Res} & \emph{\scriptsize ID} & \emph{\scriptsize Name} & \emph{\scriptsize Res}\tabularnewline
\hline 
\hline
\endhead
\hline 
\selectlanguage{english}%
\textsf{\textbf{\scriptsize 1033}}\selectlanguage{british}%
 & \textbf{\scriptsize NMR Metabolomics Investig.} & \selectlanguage{english}%
\textsf{\textbf{\scriptsize Y}}\selectlanguage{british}%
 & \selectlanguage{english}%
\textsf{\scriptsize 1362}\selectlanguage{british}%
 & {\scriptsize Hymenoptera Anatomy} & \selectlanguage{english}%
\textsf{\scriptsize N}\selectlanguage{british}%
 & \selectlanguage{english}%
\textsf{\textbf{\scriptsize 1552}}\selectlanguage{british}%
 & \textbf{\scriptsize Reprod. Trait and Phenotype} & \selectlanguage{english}%
\textsf{\textbf{\scriptsize Y}}\selectlanguage{british}%
\tabularnewline
\hline 
\selectlanguage{english}%
\textsf{\scriptsize 1039}\selectlanguage{british}%
 & {\scriptsize Proteomics Data} & \selectlanguage{english}%
\textsf{\scriptsize N}\selectlanguage{british}%
 & \selectlanguage{english}%
\textsf{\textbf{\scriptsize 1369}}\selectlanguage{british}%
 & \textbf{\scriptsize PhysicalFields} & \selectlanguage{english}%
\textsf{\textbf{\scriptsize Y}}\selectlanguage{british}%
 & \selectlanguage{english}%
\textsf{\scriptsize 1560}\selectlanguage{british}%
 & {\scriptsize Cognitive Paradigm} & \selectlanguage{english}%
\textsf{\scriptsize N}\selectlanguage{british}%
\tabularnewline
\hline 
\selectlanguage{english}%
\textsf{\scriptsize 1052}\selectlanguage{british}%
 & {\scriptsize Proteins} & \selectlanguage{english}%
\textsf{\scriptsize N}\selectlanguage{british}%
 & \selectlanguage{english}%
\textsf{\textbf{\scriptsize 1381}}\selectlanguage{british}%
 & \textbf{\scriptsize NIF Dysfunction} & \selectlanguage{english}%
\textsf{\textbf{\scriptsize Y}}\selectlanguage{british}%
 & \selectlanguage{english}%
\textsf{\textbf{\scriptsize 1565}}\selectlanguage{british}%
 & \textbf{\scriptsize Medical Social Entities} & \selectlanguage{english}%
\textsf{\textbf{\scriptsize Y}}\selectlanguage{british}%
\tabularnewline
\hline 
\selectlanguage{english}%
\textsf{\scriptsize 1054}\selectlanguage{british}%
 & {\scriptsize Amino-acid} & \selectlanguage{english}%
\textsf{\scriptsize N}\selectlanguage{british}%
 & \selectlanguage{english}%
\textsf{\scriptsize 1393}\selectlanguage{british}%
 & {\scriptsize Information Artifacts} & \selectlanguage{english}%
\textsf{\scriptsize N}\selectlanguage{british}%
 & \selectlanguage{english}%
\textsf{\textbf{\scriptsize 1567}}\selectlanguage{british}%
 & \textbf{\scriptsize Pharmacovigilance} & \selectlanguage{english}%
\textsf{\textbf{\scriptsize Y}}\selectlanguage{british}%
\tabularnewline
\hline 
\selectlanguage{english}%
\textsf{\scriptsize 1056}\selectlanguage{british}%
 & {\scriptsize Basic Vertebrate Anatomy} & \selectlanguage{english}%
\textsf{\scriptsize N}\selectlanguage{british}%
 & \selectlanguage{english}%
\textsf{\scriptsize 1394}\selectlanguage{british}%
 & {\scriptsize Syndromic Surveillance} & \selectlanguage{english}%
\textsf{\scriptsize N}\selectlanguage{british}%
 & \selectlanguage{english}%
\textsf{\scriptsize 1569}\selectlanguage{british}%
 & {\scriptsize Host Pathogen Interactions} & \selectlanguage{english}%
\textsf{\scriptsize N}\selectlanguage{british}%
\tabularnewline
\hline 
\selectlanguage{english}%
\textsf{\scriptsize 1058}\selectlanguage{british}%
 & {\scriptsize SNP} & \selectlanguage{english}%
\textsf{\scriptsize N}\selectlanguage{british}%
 & \selectlanguage{english}%
\textsf{\textbf{\scriptsize 1398}}\selectlanguage{british}%
 & \textbf{\scriptsize Language Disorder in Autism} & \selectlanguage{english}%
\textsf{\textbf{\scriptsize Y}}\selectlanguage{british}%
 & \selectlanguage{english}%
\textsf{\textbf{\scriptsize 1570}}\selectlanguage{british}%
 & \textbf{\scriptsize Traditional Med. Constitution} & \selectlanguage{english}%
\textsf{\textbf{\scriptsize Y}}\selectlanguage{british}%
\tabularnewline
\hline 
\selectlanguage{english}%
\textsf{\scriptsize 1059}\selectlanguage{british}%
 & {\scriptsize Computer-based Patient Record} & \selectlanguage{english}%
\textsf{\scriptsize N}\selectlanguage{british}%
 & \selectlanguage{english}%
\textsf{\scriptsize 1399}\selectlanguage{british}%
 & {\scriptsize Pilot Ontology} & \selectlanguage{english}%
\textsf{\scriptsize N}\selectlanguage{british}%
 & \selectlanguage{english}%
\textsf{\textbf{\scriptsize 1571}}\selectlanguage{british}%
 & \textbf{\scriptsize Traditional Med. Other Factors} & \selectlanguage{english}%
\textsf{\textbf{\scriptsize Y}}\selectlanguage{british}%
\tabularnewline
\hline 
\selectlanguage{english}%
\textsf{\scriptsize 1060}\selectlanguage{british}%
 & {\scriptsize Epoch Clinical Trial} & \selectlanguage{english}%
\textsf{\scriptsize N}\selectlanguage{british}%
 & \selectlanguage{english}%
\textsf{\scriptsize 1401}\selectlanguage{british}%
 & {\scriptsize Nursing Practice} & \selectlanguage{english}%
\textsf{\scriptsize N}\selectlanguage{british}%
 & \selectlanguage{english}%
\textsf{\textbf{\scriptsize 1572}}\selectlanguage{british}%
 & \textbf{\scriptsize Trad. Med. Signs and Symptoms} & \selectlanguage{english}%
\textsf{\textbf{\scriptsize Y}}\selectlanguage{british}%
\tabularnewline
\hline 
\selectlanguage{english}%
\textsf{\scriptsize 1061}\selectlanguage{british}%
 & {\scriptsize Pharmacogenomics} & \selectlanguage{english}%
\textsf{\scriptsize N}\selectlanguage{british}%
 & \selectlanguage{english}%
\textsf{\scriptsize 1402}\selectlanguage{british}%
 & {\scriptsize NIF Cell} & \selectlanguage{english}%
\textsf{\scriptsize N}\selectlanguage{british}%
 & \selectlanguage{english}%
\textsf{\textbf{\scriptsize 1573}}\selectlanguage{british}%
 & \textbf{\scriptsize Traditional Med. Meridian} & \selectlanguage{english}%
\textsf{\textbf{\scriptsize Y}}\selectlanguage{british}%
\tabularnewline
\hline 
\selectlanguage{english}%
\textsf{\scriptsize 1068}\selectlanguage{british}%
 & {\scriptsize Subcellular Anatomy} & \selectlanguage{english}%
\textsf{\scriptsize N}\selectlanguage{british}%
 & \selectlanguage{english}%
\textsf{\scriptsize 1406}\selectlanguage{british}%
 & {\scriptsize LinkingKing2PEP} & \selectlanguage{english}%
\textsf{\scriptsize N}\selectlanguage{british}%
 & \selectlanguage{english}%
\textsf{\scriptsize 1576}\selectlanguage{british}%
 & {\scriptsize FDA Med. Devices} & \selectlanguage{english}%
\textsf{\scriptsize N}\selectlanguage{british}%
\tabularnewline
\hline 
\selectlanguage{english}%
\textsf{\scriptsize 1082}\selectlanguage{british}%
 & {\scriptsize Gene Regulation (GRO)} & \selectlanguage{english}%
\textsf{\scriptsize N}\selectlanguage{british}%
 & \selectlanguage{english}%
\textsf{\scriptsize 1407}\selectlanguage{british}%
 & {\scriptsize Description of Dynamics} & \selectlanguage{english}%
\textsf{\scriptsize N}\selectlanguage{british}%
 & \selectlanguage{english}%
\textsf{\scriptsize 1578}\selectlanguage{british}%
 & {\scriptsize HOM-Helixhauser Scores} & \selectlanguage{english}%
\textsf{\scriptsize N}\selectlanguage{british}%
\tabularnewline
\hline 
\selectlanguage{english}%
\textsf{\scriptsize 1083}\selectlanguage{british}%
 & {\scriptsize NanoParticles} & \selectlanguage{english}%
\textsf{\scriptsize N}\selectlanguage{british}%
 & \selectlanguage{english}%
\textsf{\textbf{\scriptsize 1409}}\selectlanguage{british}%
 & \textbf{\scriptsize PKO Re} & \selectlanguage{english}%
\textsf{\textbf{\scriptsize Y}}\selectlanguage{british}%
 & \selectlanguage{english}%
\textsf{\scriptsize 1580}\selectlanguage{british}%
 & {\scriptsize Adverse Event Reporting} & \selectlanguage{english}%
\textsf{\scriptsize N}\selectlanguage{british}%
\tabularnewline
\hline 
\selectlanguage{english}%
\textsf{\textbf{\scriptsize 1084}}\selectlanguage{british}%
 & \textbf{\scriptsize NIFSTD} & \selectlanguage{english}%
\textsf{\textbf{\scriptsize Y}}\selectlanguage{british}%
 & \selectlanguage{english}%
\textsf{\scriptsize 1410}\selectlanguage{british}%
 & {\scriptsize Kinetic Simulation Algorithm} & \selectlanguage{english}%
\textsf{\scriptsize N}\selectlanguage{british}%
 & \selectlanguage{english}%
\textsf{\scriptsize 1581}\selectlanguage{british}%
 & {\scriptsize Health Indicators} & \selectlanguage{english}%
\textsf{\scriptsize N}\selectlanguage{british}%
\tabularnewline
\hline 
\selectlanguage{english}%
\textsf{\scriptsize 1086}\selectlanguage{british}%
 & {\scriptsize Disease Genetic Investigation} & \selectlanguage{english}%
\textsf{\scriptsize N}\selectlanguage{british}%
 & \selectlanguage{english}%
\textsf{\scriptsize 1411}\selectlanguage{british}%
 & {\scriptsize Functioning, Disability and Health} & \selectlanguage{english}%
\textsf{\scriptsize N}\selectlanguage{british}%
 & \selectlanguage{english}%
\textsf{\scriptsize 1582}\selectlanguage{british}%
 & {\scriptsize CAO} & \selectlanguage{english}%
\textsf{\scriptsize N}\selectlanguage{british}%
\tabularnewline
\hline 
\selectlanguage{english}%
\textsf{\textbf{\scriptsize 1087}}\selectlanguage{british}%
 & \textbf{\scriptsize Geographical Regions} & \selectlanguage{english}%
\textsf{\textbf{\scriptsize Y}}\selectlanguage{british}%
 & \selectlanguage{english}%
\textsf{\scriptsize 1413}\selectlanguage{british}%
 & {\scriptsize Software} & \selectlanguage{english}%
\textsf{\scriptsize N}\selectlanguage{british}%
 & \selectlanguage{english}%
\textsf{\scriptsize 1588}\selectlanguage{british}%
 & {\scriptsize General Purpose Datatypes} & \selectlanguage{english}%
\textsf{\scriptsize N}\selectlanguage{british}%
\tabularnewline
\hline 
\selectlanguage{english}%
\textsf{\scriptsize 1088}\selectlanguage{british}%
 & {\scriptsize MaHCO} & \selectlanguage{english}%
\textsf{\scriptsize N}\selectlanguage{british}%
 & \selectlanguage{english}%
\textsf{\textbf{\scriptsize 1414}}\selectlanguage{british}%
 & \textbf{\scriptsize General Medical Science} & \selectlanguage{english}%
\textsf{\textbf{\scriptsize Y}}\selectlanguage{british}%
 & \selectlanguage{english}%
\textsf{\scriptsize 1596}\selectlanguage{british}%
 & {\scriptsize HOM\_MDCs-DRGS} & \selectlanguage{english}%
\textsf{\scriptsize N}\selectlanguage{british}%
\tabularnewline
\hline 
\selectlanguage{english}%
\textsf{\scriptsize 1089}\selectlanguage{british}%
 & {\scriptsize BIRNLex} & \selectlanguage{english}%
\textsf{\scriptsize N}\selectlanguage{british}%
 & \selectlanguage{english}%
\textsf{\scriptsize 1415}\selectlanguage{british}%
 & {\scriptsize CTCAE} & \selectlanguage{english}%
\textsf{\scriptsize N}\selectlanguage{british}%
 & \selectlanguage{english}%
\textsf{\scriptsize 1613}\selectlanguage{british}%
 & {\scriptsize Bone Dysplasia} & \selectlanguage{english}%
\textsf{\scriptsize N}\selectlanguage{british}%
\tabularnewline
\hline 
\selectlanguage{english}%
\textsf{\scriptsize 1092}\selectlanguage{british}%
 & {\scriptsize Infectious Diseases} & \selectlanguage{english}%
\textsf{\scriptsize N}\selectlanguage{british}%
 & \selectlanguage{english}%
\textsf{\scriptsize 1417}\selectlanguage{british}%
 & {\scriptsize Influenza} & \selectlanguage{english}%
\textsf{\scriptsize N}\selectlanguage{british}%
 & \selectlanguage{english}%
\textsf{\scriptsize 1615}\selectlanguage{british}%
 & {\scriptsize Chemogenomics} & \selectlanguage{english}%
\textsf{\scriptsize N}\selectlanguage{british}%
\tabularnewline
\hline 
\selectlanguage{english}%
\textsf{\scriptsize 1100}\selectlanguage{british}%
 & {\scriptsize Genetic Intervals} & \selectlanguage{english}%
\textsf{\scriptsize N}\selectlanguage{british}%
 & \selectlanguage{english}%
\textsf{\scriptsize 1418}\selectlanguage{british}%
 & {\scriptsize TOK} & \selectlanguage{english}%
\textsf{\scriptsize N}\selectlanguage{british}%
 & \selectlanguage{english}%
\textsf{\textbf{\scriptsize 1616}}\selectlanguage{british}%
 & \textbf{\scriptsize Phylogenetics} & \selectlanguage{english}%
\textsf{\textbf{\scriptsize Y}}\selectlanguage{british}%
\tabularnewline
\hline 
\selectlanguage{english}%
\textsf{\scriptsize 1104}\selectlanguage{british}%
 & {\scriptsize Biomedical Resource} & \selectlanguage{english}%
\textsf{\scriptsize N}\selectlanguage{british}%
 & \selectlanguage{english}%
\textsf{\scriptsize 1438}\selectlanguage{british}%
 & {\scriptsize Breast tissue cell lines} & \selectlanguage{english}%
\textsf{\scriptsize N}\selectlanguage{british}%
 & \selectlanguage{english}%
\textsf{\scriptsize 1625}\selectlanguage{british}%
 & {\scriptsize HOM-ICD9PCS} & \selectlanguage{english}%
\textsf{\scriptsize N}\selectlanguage{british}%
\tabularnewline
\hline 
\selectlanguage{english}%
\textsf{\scriptsize 1106}\selectlanguage{british}%
 & {\scriptsize Gene Regulation (BOOTStrep)} & \selectlanguage{english}%
\textsf{\scriptsize N}\selectlanguage{british}%
 & \selectlanguage{english}%
\textsf{\scriptsize 1439}\selectlanguage{british}%
 & {\scriptsize General Formal (GFO)} & \selectlanguage{english}%
\textsf{\scriptsize N}\selectlanguage{british}%
 & \selectlanguage{english}%
\textsf{\scriptsize 1627}\selectlanguage{british}%
 & {\scriptsize HOMERUN Metadata} & \selectlanguage{english}%
\textsf{\scriptsize N}\selectlanguage{british}%
\tabularnewline
\hline 
\selectlanguage{english}%
\textsf{\scriptsize 1116}\selectlanguage{british}%
 & {\scriptsize Bleeding History Phenotype} & \selectlanguage{english}%
\textsf{\scriptsize N}\selectlanguage{british}%
 & \selectlanguage{english}%
\textsf{\scriptsize 1440}\selectlanguage{british}%
 & {\scriptsize General Formal (GFO-Bio)} & \selectlanguage{english}%
\textsf{\scriptsize N}\selectlanguage{british}%
 & \selectlanguage{english}%
\textsf{\scriptsize 1629}\selectlanguage{british}%
 & {\scriptsize HOM-UCARE Demographics} & \selectlanguage{english}%
\textsf{\scriptsize N}\selectlanguage{british}%
\tabularnewline
\hline 
\selectlanguage{english}%
\textsf{\scriptsize 1122}\selectlanguage{british}%
 & {\scriptsize Skin Physiology} & \selectlanguage{english}%
\textsf{\scriptsize N}\selectlanguage{british}%
 & \selectlanguage{english}%
\textsf{\scriptsize 1444}\selectlanguage{british}%
 & {\scriptsize Chemical Information} & \selectlanguage{english}%
\textsf{\scriptsize N}\selectlanguage{british}%
 & \selectlanguage{english}%
\textsf{\scriptsize 1631}\selectlanguage{british}%
 & {\scriptsize HOM-Harvard} & \selectlanguage{english}%
\textsf{\scriptsize N}\selectlanguage{british}%
\tabularnewline
\hline 
\selectlanguage{english}%
\textsf{\scriptsize 1123}\selectlanguage{british}%
 & {\scriptsize Biomedical Investigations} & \selectlanguage{english}%
\textsf{\scriptsize N}\selectlanguage{british}%
 & \selectlanguage{english}%
\textsf{\scriptsize 1461}\selectlanguage{british}%
 & {\scriptsize Translational Medicine} & \selectlanguage{english}%
\textsf{\scriptsize N}\selectlanguage{british}%
 & \selectlanguage{english}%
\textsf{\scriptsize 1633}\selectlanguage{british}%
 & {\scriptsize Cognitive Alias} & \selectlanguage{english}%
\textsf{\scriptsize N}\selectlanguage{british}%
\tabularnewline
\hline 
\selectlanguage{english}%
\textsf{\scriptsize 1126}\selectlanguage{british}%
 & {\scriptsize Family Health History} & \selectlanguage{english}%
\textsf{\scriptsize N}\selectlanguage{british}%
 & \selectlanguage{english}%
\textsf{\scriptsize 1484}\selectlanguage{british}%
 & {\scriptsize External Causes of Injuries} & \selectlanguage{english}%
\textsf{\scriptsize N}\selectlanguage{british}%
 & \selectlanguage{english}%
\textsf{\textbf{\scriptsize 1638}}\selectlanguage{british}%
 & \textbf{\scriptsize Data Mining} & \selectlanguage{english}%
\textsf{\textbf{\scriptsize Y}}\selectlanguage{british}%
\tabularnewline
\hline 
\selectlanguage{english}%
\textsf{\scriptsize 1128}\selectlanguage{british}%
 & {\scriptsize Comparative Data Analysis} & \selectlanguage{english}%
\textsf{\scriptsize N}\selectlanguage{british}%
 & \selectlanguage{english}%
\textsf{\textbf{\scriptsize 1487}}\selectlanguage{british}%
 & \textbf{\scriptsize Body System} & \selectlanguage{english}%
\textsf{\textbf{\scriptsize Y}}\selectlanguage{british}%
 & \selectlanguage{english}%
\textsf{\scriptsize 1639}\selectlanguage{british}%
 & {\scriptsize Epilepsy} & \selectlanguage{english}%
\textsf{\scriptsize N}\selectlanguage{british}%
\tabularnewline
\hline 
\selectlanguage{english}%
\textsf{\scriptsize 1130}\selectlanguage{british}%
 & {\scriptsize Cancer Research and Mgmt} & \selectlanguage{english}%
\textsf{\scriptsize N}\selectlanguage{british}%
 & \selectlanguage{english}%
\textsf{\scriptsize 1488}\selectlanguage{british}%
 & {\scriptsize SysMO-JERM} & \selectlanguage{english}%
\textsf{\scriptsize N}\selectlanguage{british}%
 & \selectlanguage{english}%
\textsf{\scriptsize 1640}\selectlanguage{british}%
 & {\scriptsize Pediatric Terminology} & \selectlanguage{english}%
\textsf{\scriptsize N}\selectlanguage{british}%
\tabularnewline
\hline 
\selectlanguage{english}%
\textsf{\scriptsize 1131}\selectlanguage{british}%
 & {\scriptsize MGED} & \selectlanguage{english}%
\textsf{\scriptsize N}\selectlanguage{british}%
 & \selectlanguage{english}%
\textsf{\scriptsize 1489}\selectlanguage{british}%
 & {\scriptsize Adverse Events} & \selectlanguage{english}%
\textsf{\scriptsize N}\selectlanguage{british}%
 & \selectlanguage{english}%
\textsf{\scriptsize 1641}\selectlanguage{british}%
 & {\scriptsize HOM-ICD9CM-ECODES} & \selectlanguage{english}%
\textsf{\scriptsize N}\selectlanguage{british}%
\tabularnewline
\hline 
\selectlanguage{english}%
\textsf{\scriptsize 1134}\selectlanguage{british}%
 & {\scriptsize BioTop} & \selectlanguage{english}%
\textsf{\scriptsize N}\selectlanguage{british}%
 & \selectlanguage{english}%
\textsf{\scriptsize 1494}\selectlanguage{british}%
 & {\scriptsize Tissue Microarray} & \selectlanguage{english}%
\textsf{\scriptsize N}\selectlanguage{british}%
 & \selectlanguage{english}%
\textsf{\scriptsize 1642}\selectlanguage{british}%
 & {\scriptsize HOM-DXPROCS MDCDRG} & \selectlanguage{english}%
\textsf{\scriptsize N}\selectlanguage{british}%
\tabularnewline
\hline 
\selectlanguage{english}%
\textsf{\textbf{\scriptsize 1136}}\selectlanguage{british}%
 & \textbf{\scriptsize Experimental Factors} & \selectlanguage{english}%
\textsf{\textbf{\scriptsize Y}}\selectlanguage{british}%
 & \selectlanguage{english}%
\textsf{\scriptsize 1497}\selectlanguage{british}%
 & {\scriptsize PMA 2010} & \selectlanguage{english}%
\textsf{\scriptsize N}\selectlanguage{british}%
 & \selectlanguage{english}%
\textsf{\scriptsize 1643}\selectlanguage{british}%
 & {\scriptsize HOM-ICD9\_PROCS OSHPD} & \selectlanguage{english}%
\textsf{\scriptsize N}\selectlanguage{british}%
\tabularnewline
\hline 
\selectlanguage{english}%
\textsf{\scriptsize 1141}\selectlanguage{british}%
 & {\scriptsize Physics for Biology} & \selectlanguage{english}%
\textsf{\scriptsize N}\selectlanguage{british}%
 & \selectlanguage{english}%
\textsf{\scriptsize 1500}\selectlanguage{british}%
 & {\scriptsize RNA} & \selectlanguage{english}%
\textsf{\scriptsize N}\selectlanguage{british}%
 & \selectlanguage{english}%
\textsf{\scriptsize 1648}\selectlanguage{british}%
 & {\scriptsize HOM-DATASOURCE OSHPD} & \selectlanguage{english}%
\textsf{\scriptsize N}\selectlanguage{british}%
\tabularnewline
\hline 
\selectlanguage{english}%
\textsf{\textbf{\scriptsize 1142}}\selectlanguage{british}%
 & \textbf{\scriptsize Cardiac Electrophysiology} & \selectlanguage{english}%
\textsf{\textbf{\scriptsize Y}}\selectlanguage{british}%
 & \selectlanguage{english}%
\textsf{\scriptsize 1501}\selectlanguage{british}%
 & {\scriptsize Neomark Oral Cancer-based} & \selectlanguage{english}%
\textsf{\scriptsize N}\selectlanguage{british}%
 & \selectlanguage{english}%
\textsf{\textbf{\scriptsize 1650}}\selectlanguage{british}%
 & \textbf{\scriptsize Units} & \selectlanguage{english}%
\textsf{\textbf{\scriptsize Y}}\selectlanguage{british}%
\tabularnewline
\hline 
\selectlanguage{english}%
\textsf{\scriptsize 1146}\selectlanguage{british}%
 & {\scriptsize Electrocardiography} & \selectlanguage{english}%
\textsf{\scriptsize N}\selectlanguage{british}%
 & \selectlanguage{english}%
\textsf{\scriptsize 1505}\selectlanguage{british}%
 & {\scriptsize MicroRNA Target Prediction} & \selectlanguage{english}%
\textsf{\scriptsize N}\selectlanguage{british}%
 & \selectlanguage{english}%
\textsf{\scriptsize 1652}\selectlanguage{british}%
 & {\scriptsize HOM-OSHPD Use cases} & \selectlanguage{english}%
\textsf{\scriptsize N}\selectlanguage{british}%
\tabularnewline
\hline 
\selectlanguage{english}%
\textsf{\scriptsize 1149}\selectlanguage{british}%
 & {\scriptsize Dermatology Lexicon} & \selectlanguage{english}%
\textsf{\scriptsize N}\selectlanguage{british}%
 & \selectlanguage{english}%
\textsf{\textbf{\scriptsize 1515}}\selectlanguage{british}%
 & \textbf{\scriptsize Interaction Network} & \selectlanguage{english}%
\textsf{\textbf{\scriptsize Y}}\selectlanguage{british}%
 & \selectlanguage{english}%
\textsf{\scriptsize 1653}\selectlanguage{british}%
 & {\scriptsize HOM-PROCS2 OSHPD} & \selectlanguage{english}%
\textsf{\scriptsize N}\selectlanguage{british}%
\tabularnewline
\hline 
\selectlanguage{english}%
\textsf{\scriptsize 1172}\selectlanguage{british}%
 & {\scriptsize Vaccines} & \selectlanguage{english}%
\textsf{\scriptsize N}\selectlanguage{british}%
 & \selectlanguage{english}%
\textsf{\textbf{\scriptsize 1521}}\selectlanguage{british}%
 & \textbf{\scriptsize Neural Motor Recovery} & \selectlanguage{english}%
\textsf{\textbf{\scriptsize Y}}\selectlanguage{british}%
 & \selectlanguage{english}%
\textsf{\scriptsize 1658}\selectlanguage{british}%
 & {\scriptsize Hewan Invertebrata} & \selectlanguage{english}%
\textsf{\scriptsize N}\selectlanguage{british}%
\tabularnewline
\hline 
\selectlanguage{english}%
\textsf{\scriptsize 1183}\selectlanguage{british}%
 & {\scriptsize Lipids} & \selectlanguage{english}%
\textsf{\scriptsize N}\selectlanguage{british}%
 & \selectlanguage{english}%
\textsf{\scriptsize 1522}\selectlanguage{british}%
 & {\scriptsize BioPAX} & \selectlanguage{english}%
\textsf{\scriptsize N}\selectlanguage{british}%
 & \selectlanguage{english}%
\textsf{\scriptsize 1665}\selectlanguage{british}%
 & {\scriptsize Student Health Record} & \selectlanguage{english}%
\textsf{\scriptsize N}\selectlanguage{british}%
\tabularnewline
\hline 
\selectlanguage{english}%
\textsf{\scriptsize 1190}\selectlanguage{british}%
 & {\scriptsize Parasite LifeCycle} & \selectlanguage{english}%
\textsf{\scriptsize N}\selectlanguage{british}%
 & \selectlanguage{english}%
\textsf{\scriptsize 1523}\selectlanguage{british}%
 & {\scriptsize OBOE-SBC} & \selectlanguage{english}%
\textsf{\scriptsize N}\selectlanguage{british}%
 & \selectlanguage{english}%
\textsf{\scriptsize 1666}\selectlanguage{british}%
 & {\scriptsize Emotion} & \selectlanguage{english}%
\textsf{\scriptsize N}\selectlanguage{british}%
\tabularnewline
\hline 
\selectlanguage{english}%
\textsf{\scriptsize 1192}\selectlanguage{british}%
 & {\scriptsize Proteomics Pipeline Infrastructures} & \selectlanguage{english}%
\textsf{\scriptsize N}\selectlanguage{british}%
 & \selectlanguage{english}%
\textsf{\scriptsize 1524}\selectlanguage{british}%
 & {\scriptsize OBOE} & \selectlanguage{english}%
\textsf{\scriptsize N}\selectlanguage{british}%
 & \selectlanguage{english}%
\textsf{\scriptsize 1667}\selectlanguage{british}%
 & {\scriptsize HOM-DATASOURCE OSHPDSC} & \selectlanguage{english}%
\textsf{\scriptsize N}\selectlanguage{british}%
\tabularnewline
\hline 
\selectlanguage{english}%
\textsf{\scriptsize 1237}\selectlanguage{british}%
 & {\scriptsize Situation-based Access Control} & \selectlanguage{english}%
\textsf{\scriptsize N}\selectlanguage{british}%
 & \selectlanguage{english}%
\textsf{\scriptsize 1530}\selectlanguage{british}%
 & {\scriptsize Animal natural history} & \selectlanguage{english}%
\textsf{\scriptsize N}\selectlanguage{british}%
 & \selectlanguage{english}%
\textsf{\scriptsize 1668}\selectlanguage{british}%
 & {\scriptsize HOM-OSHPD-SC} & \selectlanguage{english}%
\textsf{\scriptsize N}\selectlanguage{british}%
\tabularnewline
\hline 
\selectlanguage{english}%
\textsf{\scriptsize 1247}\selectlanguage{british}%
 & {\scriptsize GeoSpecies} & \selectlanguage{english}%
\textsf{\scriptsize N}\selectlanguage{british}%
 & \selectlanguage{english}%
\textsf{\textbf{\scriptsize 1532}}\selectlanguage{british}%
 & \textbf{\scriptsize SemanticScience} & \selectlanguage{english}%
\textsf{\textbf{\scriptsize Y}}\selectlanguage{british}%
 & \selectlanguage{english}%
\textsf{\scriptsize 1671}\selectlanguage{british}%
 & {\scriptsize Quantitative Imaging Biomarkers} & \selectlanguage{english}%
\textsf{\scriptsize N}\selectlanguage{british}%
\tabularnewline
\hline 
\selectlanguage{english}%
\textsf{\scriptsize 1249}\selectlanguage{british}%
 & {\scriptsize Smoking Behavior Risk} & \selectlanguage{english}%
\textsf{\scriptsize N}\selectlanguage{british}%
 & \selectlanguage{english}%
\textsf{\scriptsize 1533}\selectlanguage{british}%
 & {\scriptsize BioAssay} & \selectlanguage{english}%
\textsf{\scriptsize N}\selectlanguage{british}%
 & \selectlanguage{english}%
\textsf{\scriptsize 1672}\selectlanguage{british}%
 & {\scriptsize DIKB-Evidence} & \selectlanguage{english}%
\textsf{\scriptsize N}\selectlanguage{british}%
\tabularnewline
\hline 
\selectlanguage{english}%
\textsf{\textbf{\scriptsize 1290}}\selectlanguage{british}%
 & \textbf{\scriptsize ABA Adult Mouse Brain} & \selectlanguage{english}%
\textsf{\textbf{\scriptsize Y}}\selectlanguage{british}%
 & \selectlanguage{english}%
\textsf{\textbf{\scriptsize 1534}}\selectlanguage{british}%
 & \textbf{\scriptsize Apollo-akesios} & \selectlanguage{english}%
\textsf{\textbf{\scriptsize Y}}\selectlanguage{british}%
 & \selectlanguage{english}%
\textsf{\scriptsize 1676}\selectlanguage{british}%
 & {\scriptsize Randomized Controlled Trials} & \selectlanguage{english}%
\textsf{\scriptsize N}\selectlanguage{british}%
\tabularnewline
\hline 
\selectlanguage{english}%
\textsf{\scriptsize 1304}\selectlanguage{british}%
 & {\scriptsize Breast Cancer Grading} & \selectlanguage{english}%
\textsf{\scriptsize N}\selectlanguage{british}%
 & \selectlanguage{english}%
\textsf{\scriptsize 1537}\selectlanguage{british}%
 & {\scriptsize Brucellosis} & \selectlanguage{english}%
\textsf{\scriptsize N}\selectlanguage{british}%
 & \selectlanguage{english}%
\textsf{\textbf{\scriptsize 1686}}\selectlanguage{british}%
 & \textbf{\scriptsize Neomark Oral Cancer} & \selectlanguage{english}%
\textsf{\textbf{\scriptsize Y}}\selectlanguage{british}%
\tabularnewline
\hline 
\selectlanguage{english}%
\textsf{\scriptsize 1314}\selectlanguage{british}%
 & {\scriptsize Cell Line (2)} & \selectlanguage{english}%
\textsf{\scriptsize N}\selectlanguage{british}%
 & \selectlanguage{english}%
\textsf{\textbf{\scriptsize 1538}}\selectlanguage{british}%
 & \textbf{\scriptsize Roles} & \selectlanguage{english}%
\textsf{\textbf{\scriptsize Y}}\selectlanguage{british}%
 & \selectlanguage{english}%
\textsf{\textbf{\scriptsize 1696}}\selectlanguage{british}%
 & \textbf{\scriptsize Synapses} & \selectlanguage{english}%
\textsf{\textbf{\scriptsize Y}}\selectlanguage{british}%
\tabularnewline
\hline 
\selectlanguage{english}%
\textsf{\scriptsize 1321}\selectlanguage{british}%
 & {\scriptsize Neural ElectroMagnetic Ontology} & \selectlanguage{english}%
\textsf{\scriptsize N}\selectlanguage{british}%
 & \selectlanguage{english}%
\textsf{\scriptsize 1540}\selectlanguage{british}%
 & {\scriptsize Drug Discovery Investigations} & \selectlanguage{english}%
\textsf{\scriptsize N}\selectlanguage{british}%
 & \selectlanguage{english}%
\textsf{\textbf{\scriptsize 3002}}\selectlanguage{british}%
 & \textbf{\scriptsize Mental Functioning} & \selectlanguage{english}%
\textsf{\textbf{\scriptsize Y}}\selectlanguage{british}%
\tabularnewline
\hline 
\selectlanguage{english}%
\textsf{\textbf{\scriptsize 1332}}\selectlanguage{british}%
 & \textbf{\scriptsize Basic Formal Ontology} & \selectlanguage{english}%
\textsf{\textbf{\scriptsize Y}}\selectlanguage{british}%
 & \selectlanguage{english}%
\textsf{\scriptsize 1541}\selectlanguage{british}%
 & {\scriptsize Cell Line (MCCL 2)} & \selectlanguage{english}%
\textsf{\scriptsize N}\selectlanguage{british}%
 & \multicolumn{1}{c}{} & \multicolumn{1}{c}{} & \multicolumn{1}{c}{}\tabularnewline
\cline{1-6} 
\selectlanguage{english}%
\textsf{\scriptsize 1335}\selectlanguage{british}%
 & {\scriptsize Parasite Experiments} & \selectlanguage{english}%
\textsf{\scriptsize N}\selectlanguage{british}%
 & \selectlanguage{english}%
\textsf{\scriptsize 1550}\selectlanguage{british}%
 & {\scriptsize PHARE} & \selectlanguage{english}%
\textsf{\scriptsize N}\selectlanguage{british}%
 & \multicolumn{1}{c}{} & \multicolumn{1}{c}{} & \multicolumn{1}{c}{}\tabularnewline
\cline{1-6} 
\end{longtable}
\end{document}